\documentclass{MITcsail}

\usepackage{amsmath,amsfonts,bm}

\def\eqref#1{equation~\ref{#1}}

\def\1{\bm{1}}

\DeclareMathAlphabet{\mathsfit}{\encodingdefault}{\sfdefault}{m}{sl}
\SetMathAlphabet{\mathsfit}{bold}{\encodingdefault}{\sfdefault}{bx}{n}

\definecolor{darkred}{RGB}{140, 21, 21}
\definecolor{lightgray}{gray}{0.5}
\definecolor{orange}{HTML}{F58025}
\definecolor{deepred}{rgb}{0.631,0.102,0.102}
\definecolor{amethyst}{rgb}{0.6, 0.4, 0.8}
\definecolor{darkgreen}{rgb}{0.3,0.7,0.3}
\definecolor{salmon}{RGB}{241, 150, 141}
\definecolor{mildyellow}{HTML}{FFF2CC}
\definecolor{jhulightblue}{HTML}{68ACE5}

\usepackage{hyperref}
\usepackage{xcolor}
\hypersetup{
    colorlinks=true,
    linkcolor=jhulightblue,
    citecolor=lightgray,
    filecolor=darkred,
    urlcolor=jhulightblue
}
\usepackage[authoryear]{natbib}
\usepackage{amsmath}
\usepackage{amsfonts}
\usepackage{amssymb}
\usepackage{wrapfig}
\usepackage{subcaption}
\usepackage{multirow}
 \usepackage{mathtools} 

\usepackage{verbatim}

\usepackage{anyfontsize}

\usepackage{microtype}
\usepackage{graphicx}
\usepackage{booktabs} 
\usepackage{multirow}
\usepackage{amsmath,amssymb}
\usepackage{booktabs}
\usepackage{caption,subcaption}
\usepackage{svg}

\usepackage{authblk}

\renewcommand\Affilfont{\normalsize\normalfont}

\definecolor{mygreen}{HTML}{3cb44b}
\definecolor{skyblue}{HTML}{beffff}
\definecolor{lightgreen}{HTML}{90ee90}

\usepackage{color, colortbl}

\definecolor{emerald}{rgb}{0.31, 0.78, 0.37}

\usepackage{datetime}
\newdateformat{ymd}{\the\year-\the\month-\the\day}

\usepackage{tcolorbox}
\usepackage{enumitem}
\definecolor{mygreen}{HTML}{3cb44b}
\colorlet{myyellow}{green!10!orange!90!}
\makeatletter

\usepackage{tikz}
\usetikzlibrary{arrows,shapes,snakes,automata,backgrounds,fit,petri}
\usepackage{adjustbox}

\newcommand{\RN}[1]{%
	\textup{\lowercase\expandafter{\it \romannumeral#1}}%
}
\usepackage{tabu}

\newcommand{\beq}{\vspace{0mm}\begin{equation}}
\newcommand{\eeq}{\vspace{0mm}\end{equation}}
\newcommand{\beqs}{\vspace{0mm}\begin{eqnarray}}
\newcommand{\eeqs}{\vspace{0mm}\end{eqnarray}}
\newcommand{\barr}{\begin{array}}
\newcommand{\earr}{\end{array}}

\usepackage{color, colortbl}
\definecolor{Gray}{gray}{0.93}

\usepackage{lipsum}

\usepackage{pifont}
\newcommand{\cmark}{\ding{51}}%
\newcommand{\xmark}{\ding{55}}%

\usepackage{makecell}

\usepackage{xcolor,amsmath}
\usepackage[ruled,vlined,linesnumbered]{algorithm2e}
\SetKwComment{Comment}{$\triangleright$ }{}
\SetKwInput{KwInput}{Input}
\SetKwInput{KwOutput}{Output}
\DontPrintSemicolon

\usepackage{xcolor}
\definecolor{mygreen}{HTML}{3cb44b}

\SetKwComment{Comment}{\color{green!50!black}\# }{}

\SetKwProg{Function}{def}{:}{}

\SetKwProg{For}{for}{:}{}
\SetKwProg{If}{if}{:}{}

\usepackage{microtype}
\usepackage{listings}
\usepackage{float}
\usepackage{graphicx}
\usepackage{mathtools}
\usepackage{bbding}
\usepackage{enumitem}
\usepackage{siunitx}
\usepackage[table]{xcolor}
\usepackage{tabularx}
\usepackage{array}
\usepackage[capitalize,noabbrev]{cleveref}

\newenvironment{addmargin}[2][]{%
  \begin{list}{}{\setlength{\leftmargin}{#1}\setlength{\rightmargin}{#2}}\item[]%
}{\end{list}}

\newcolumntype{Y}{>{\centering\arraybackslash}X}
\newcolumntype{R}{>{\raggedleft\arraybackslash}X}

\definecolor{gred}{rgb}{0.859,0.267,0.216}
\definecolor{ggreen}{rgb}{0.059,0.616,0.345}
\AtBeginDocument{%
  \renewcommand{\cmark}{\textcolor{ggreen}{\ding{51}}}%
  \renewcommand{\xmark}{\textcolor{gred}{\ding{55}}}%
}

\theoremstyle{plain}

\theoremstyle{definition}

\theoremstyle{remark}

\newcommand{\model}{}
\newcommand{\pagehead}{\model}

\title{3D-Belief: Embodied Belief Inference via Generative 3D World Modeling}

\author{Yifan Yin\textsuperscript{*}, Zehao Wen\textsuperscript{*}, Suyu Ye, Jieneng Chen, Zehan Zheng, Nanru Dai, Haojun Shi,
Aydan Huang, Zheyuan Zhang, Alan Yuille, Jianwen Xie\textsuperscript{$\dagger$}, Ayush Tewari\textsuperscript{$\ddagger$}, Tianmin Shu \\
\vspace{0.6em}
\normalfont{Johns Hopkins University \quad $^{\dagger}$Lambda \quad $^{\ddagger}$University of Cambridge} \\
\vspace{0.6em}
\texttt{\href{https://3d-belief.github.io/}{ 3d-belief.github.io}}
}

\begin{document}

\maketitle
\thispagestyle{firstpagestyle}

\begin{abstract}
\textbf{\textit{Abstract}}. Recent advances in visual generative models have highlighted the promise of learning generative world models. However, most existing approaches frame world modeling as novel-view synthesis or future-frame prediction, emphasizing visual realism rather than the structured uncertainty required by embodied agents acting under partial observability. In this work, we propose a different perspective: world modeling as embodied belief inference in 3D space. From this view, a world model should not merely render what may be seen, but maintain and update an agent’s belief about the unobserved 3D world as new observations are acquired. We identify several key capabilities for such models, including spatially consistent scene memory, multi-hypothesis belief sampling, sequential belief updating, and semantically informed prediction of unseen regions. We instantiate these ideas in 3D-Belief, a generative 3D world model that infers explicit, actionable 3D beliefs from partial observations and updates them online over time. Unlike prior visual prediction models, 3D-Belief represents uncertainty directly in 3D, enabling embodied agents to imagine plausible scene completions and reason over partially observed environments. We evaluate 3D-Belief on 2D visual quality for scene memory and unobserved-scene imagination, object- and scene-level 3D imagination using our proposed 3D-CORE benchmark, and challenging object navigation tasks in both simulation and the real world. Experiments show that 3D-Belief improves 2D and 3D imagination quality and downstream embodied task performance compared to state-of-the-art methods.\footnote{Equal contribution. Videos and demos are available at \url{https://3d-belief.github.io/}.}
\end{abstract}

\section{Introduction} \label{intro}

Recent advances in video generation models have shown promising results on learning a generative world model that can predict future frames based on a single, multi-view, or streaming image inputs \cite{song2025history, bar2025navigation, ren2025gen3c, yu2024viewcrafter, yu2025trajectorycrafter, gu2025diffusion, ma2025you, wu2025difix3d+, huang2025voyager, cao2025uni3c}. These models have demonstrated the remarkable ability to render novel views of unobserved parts of the environments or the change of the seen parts of the environments caused by agents' actions.

However, there remains a substantial gap between what these models are trained to do and what a world model must represent for embodied agents operating from partial observations of a 3D scene. Predicting pixels is not the same as inferring the underlying 3D world: an agent in an unfamiliar environment must form a belief over both observed and unobserved regions, where this belief should capture task-relevant structure rather than only visual appearance. As new observations arrive, the agent should update this belief accordingly, forming an ever-evolving representation of the 3D world.

\begin{figure*}[t]
\centering
\includegraphics[width=1.0\textwidth]{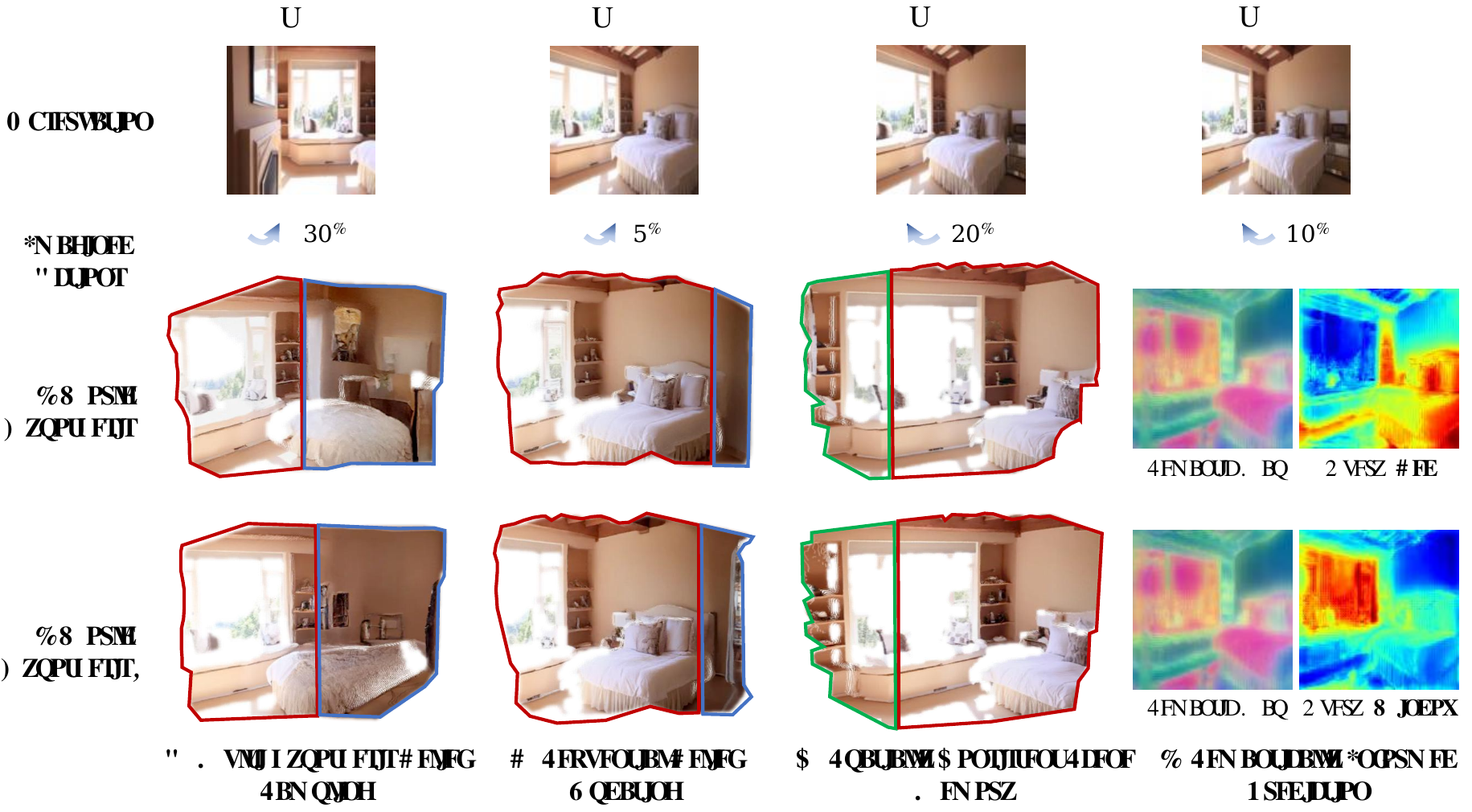}
\caption{\textbf{Key capabilities of a generative 3D belief model.} An agent performs sequential mental simulation in an unknown environment. Top: selected RGB observations; second row: imagined actions; bottom: two sampled 3D world hypotheses. Regions supported by current observations are marked in \textcolor{red}{red}, imagined unseen regions in \textcolor{blue}{blue}, and recalled content from scene memory in \textcolor{ggreen}{green}. \textbf{A. Multi-hypothesis belief sampling:} at $t{=}0$, multiple plausible completions exist for unseen structure (e.g., unseen part of the bed and room). \textbf{B. Sequential belief updating:} incorporating observations from $t{=}0$ to $105$ resolves ambiguity and updates the hypotheses (e.g., the appearance of the bed). \textbf{C. Spatially consistent scene memory:} revisiting viewpoints in mental simulation retrieves previously seen structure (e.g., bookshelf) with object permanence. \textbf{D. Semantically informed prediction:} semantic maps support task queries (e.g., \emph{bed}, \emph{window}).}\label{fig:intro}
\end{figure*}

To achieve this, we theorize that robust embodied agents require a generative world model that has the following key capacities, as illustrated in Figure~\ref{fig:intro}. First, the model needs to have a \textbf{spatially consistent scene memory}, which preserves the geometry and semantics of observed parts of the 3D world. Second, the model output should allow \textbf{multi-hypothesis belief sampling} to model uncertainty over the unseen parts of the 3D world. Third, the sequent nature of embodied tasks requires \textbf{sequential belief updating} conditioned on streaming partial observations. Lastly, it is crucial to produce \textbf{semantically informed belief prediction} which provides direct semantic prediction of the unseen parts of the 3D world to drive reasoning and planning. 

As summarized in Table~\ref{tab:relatedwork-compare} in Appendix~\ref{appendix:related_work}, existing methods cover only a subset of these capabilities. Pose- or action-conditioned video generation models~\cite{song2025history, wu2025geometry, bar2025navigation} provide strong 2D imagination from streaming observations, but remain in pixel space without explicit, actionable 3D beliefs. Methods that augment visual generation with 3D caches~\cite{ren2025gen3c, yu2024viewcrafter, yu2025trajectorycrafter, zhou2025learning} enable sequential scene memory, but mainly use 3D as a reconstruction cache rather than for diverse 3D imagination. Feedforward novel-view synthesis, reconstruction, and 3D foundation models~\cite{ye2024no, charatan2024pixelsplat, chen2024mvsplat, jiang2025anysplat, wang2024dust3r, wang2025vggt, zhuo2025streaming, wang2025continuous} maintain explicit 3D representations and may support online updates or limited inference of unseen structure, but lack uncertainty-aware multi-hypothesis 3D imagination. DFM~\cite{tewari2023diffusion} supports diverse, view-consistent imagination, but uses an implicit NeRF-style representation that is difficult to convert into explicit 3D structure for embodied tasks. Recent generative 3D world models~\cite{shen2026lyra, wang2026rein3d, li2025flashworld} improve 3D-consistent generation, but are not designed for online belief inference from streaming egocentric observations. Finally, semantic grounding remains limited: while semantic 3D scene representations have been explored~\cite{ye2026semgs}, existing generative world models still lack open-vocabulary semantic predictions for unobserved 3D regions.

To bridge the gap, we propose \textit{\textbf{3D-Belief}}, a generative 3D world model that captures all the key abilities we hypothesized above. Built upon a diffusion model, 3D-Belief learns to predict explicit 3D representations of the full world state (including unseen regions) based on past egocentric visual observations. Critically, 3D-Belief can sequentially update its 3D prediction of the full world state based on new observations while maintaining consistency with all historical observations. The resulting 3D representations contain geometric, spatial, and semantic information about the 3D world, forming a coherent and queryable belief state that can support downstream embodied tasks.

To evaluate 3D-Belief, we assess its ability to predict and update 3D scene beliefs under partial observability. We first evaluate the 2D visual quality of the belief prediction, measuring both scene memorization in observed regions and imagination of unobserved regions from egocentric observations. We then introduce a novel benchmark, 3D-CORE (3D COntextual REasoning), to probe object- and scene-level 3D imagination. The proposed reasoning tasks provide a diagnostic evaluation of the four key capabilities of embodied agents' beliefs in 3D space. Finally, we evaluate whether the learned 3D scene beliefs can support downstream embodied tasks. In particular, we focus on open-vocabulary object navigation, where a robot is instructed to search for a target object in unseen environments. We test the model in both simulation and the real world with a mobile manipulator. Across these evaluations, 3D-Belief consistently improves 2D visual quality, explicit 3D imagination, and downstream embodied task performance compared to strong baselines.

In sum, our main contributions are: (1) a conceptual framework that models embodied belief inference via generative 3D world modeling; (2) 3D-Belief, a generative 3D world model that predicts actionable, semantic, and uncertainty-aware 3D scene beliefs from partial observations; (3) 3D-CORE, a benchmark for evaluating object- and scene-level 3D imagination in embodied settings; and (4) comprehensive evaluations demonstrating the effectiveness of 3D-Belief on 2D visual quality for scene memorization and imagination, 3D scene imagination, and downstream object navigation tasks in both simulation and the real world.

\section{Related Work}

\textbf{Visual Generative Models.}
Recent video diffusion and view-synthesis models can generate realistic, high-resolution videos with controllable camera motion \cite{song2025history, huang2025voyager, yu2024viewcrafter, yu2025trajectorycrafter}. There has also been extensions to 3D/4D generation \cite{zhen2025tesseract}. However, these models primarily generate frames in pixel space and typically require separate reconstruction modules to recover an explicit 3D scene representation. In contrast, we learn a generative model that directly predicts an explicit, actionable 3D belief.

\textbf{World Model-Based Planning.}
Video world models have been used for action-conditioned rollouts to enable world model-based planning \cite{du2023video, du2024video, zhou2024robodreamer, zhang2024combo, bar2025navigation, hafner2025training}. Unlike these approaches, which plan over 2D rollouts, we show that a \emph{3D} generative world model, such as 3D-Belief, can better support planning for embodied tasks in unseen 3D environments such as open-vocabulary object navigation.

\textbf{Semantically-Embedded 3D Scene Representations.}
There has been work that augments 3D representations with semantics for open-vocabulary querying and task-directed reasoning. LERF associates text-aligned features with radiance fields for free-form language grounding \cite{kerr2023lerf}, while ConceptGraph and DynaMem build persistent, online-updated spatial-semantic memories \cite{gu2024conceptgraphs, liu2025dynamem}. These methods largely focus on representing what has been observed. In contrast, we introduce a semantically informed 3D belief that \emph{predicts} unobserved regions and supports sequential updates.

\section{3D-Belief}\label{sec:method}

\begin{figure}[t!]
  \centering
    \includegraphics[width=0.8\linewidth]{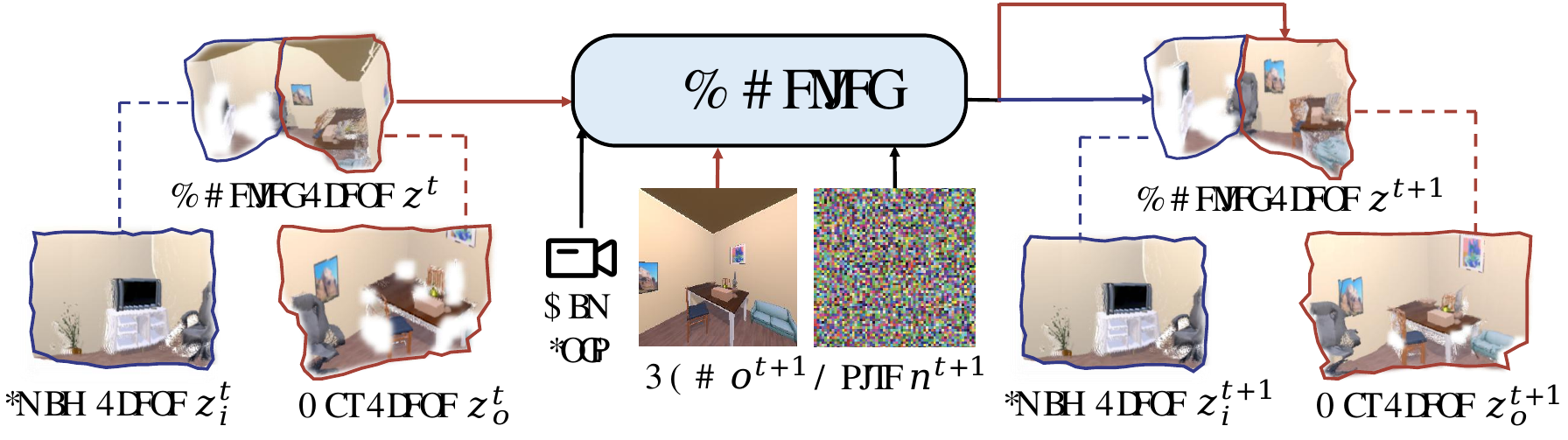}
    
  \vspace{-5pt}
  \caption{Overview of 3D-Belief. We represent the 3D representation of a sampled scene at step $t$ as $z^t$, which includes the observed and imagined Gaussians $z^t_o$ and $z^t_i$. Given $z^t$, a new observations $o^{t+1}$, and a sampled noise image $n^{t+1}$, 3D-Belief samples a new 3D scene representation at step $t+1$, i.e., $z^{t+1}$, to update the belief of the 3D world.}
  \label{fig:architecture_top}
  \vspace{-10pt}
\end{figure}

In this work, we propose a generative 3D world model, 3D-Belief. As shown in Figure~\ref{fig:architecture_top}, it learns to predict and sequentially update explicit 3D representations of a scene online. We first formalize the 3D-Belief model in Section~\ref{sec:formulation} and then introduce a model architecture to implement 3D-Belief in Section~\ref{sec:architecture} - \ref{sec:training_objective}.

\subsection{Formulation}\label{sec:formulation}

We follow the definition of the belief in partially observable Markov decision making (POMDP) \cite{kaelbling1998planning}. Specifically, belief $b(s^t)$ is a distribution of the state $s^t$ at any given step $t$ conditioned on past observations $o^t$, i.e., $b(s^t) = P(s^t | o^{1:t})$. Given new observations $o^{t+1}$ at the recent step $t+1$, we can update the belief as 
\begin{equation} \label{eq:belief}
    b(s^{t+1}) = \sum_{s^t} P(s^{t+1} | o^{t+1}, s^t)b(s^t).
\end{equation}
Directly predicting the full 3D scene for the first (3D memory) and fourth (semantics) key capabilities can be extremely challenging. Therefore, we can instead predict a 3D representation of the scene, $z^t = \phi(s^t)$. Such representations (1) should preserve the important 3D structures and semantic information of the scene necessary for downstream embodied tasks, and (2) can be parametrized so that they can be conveniently constructed via model training. In this work, we consider 3D Gaussian Splatting (3DGS) \cite{kerbl20233d}, as it not only provides an explicit 3D representation of a scene and can carry semantic feature embeddings in each primitive \cite{qiu2024feature}, but also allows fast rendering to support downstream embodied tasks. 

Formally, $z^t=\{g_k\}_{k=1}^K$ with Gaussian primitives
$g_k=(\mu_k,\Sigma_k,\alpha_k,S_k,e_k)$ (mean, covariance, opacity, SH appearance, semantic embedding).
We split $z^t=z_o^t\cup z_i^t$ into observed and imagined Gaussians. For sequential belief
updates, we rewrite~\eqref{eq:belief} as an autoregressive update:
\begin{equation}\label{eq:sequantial}
z^{t+1}\sim p(z^{t+1}\mid o^{t+1},z_o^t).
\end{equation}
In particular, $z_o^t$ is expanded to form $z_o^{t+1}$ by incorporating the new observation, whereas $z_i^t$ is replaced with a new imagination, $z_i^{t+1}$, since the previously imagined content may conflict with new evidence. This formulation (i) enables a simple training procedure that requires supervision only over short horizons (as discussed in Sec.~\ref{sec:architecture}, we train our model using image pairs), (ii) supports test-time scaling for long-horizon planning via continual belief updates, and (iii) maintains constant per-step computational cost independent of the planning horizon.

Given $z^t$, we render an imagined view from any pose $\theta$ as $\hat{o}=\mathcal{G}(z^t,\theta)$ \cite{kerbl20233d}, producing RGB, depth, and a semantic feature map from per-primitive embeddings. These renderings enable model-based mental simulation, and language queries over the semantic map let a planner score and select plans.

\subsection{Model Architecture}\label{sec:architecture}

\begin{figure}[t!]
    \centering
    \includegraphics[width=0.65\linewidth]{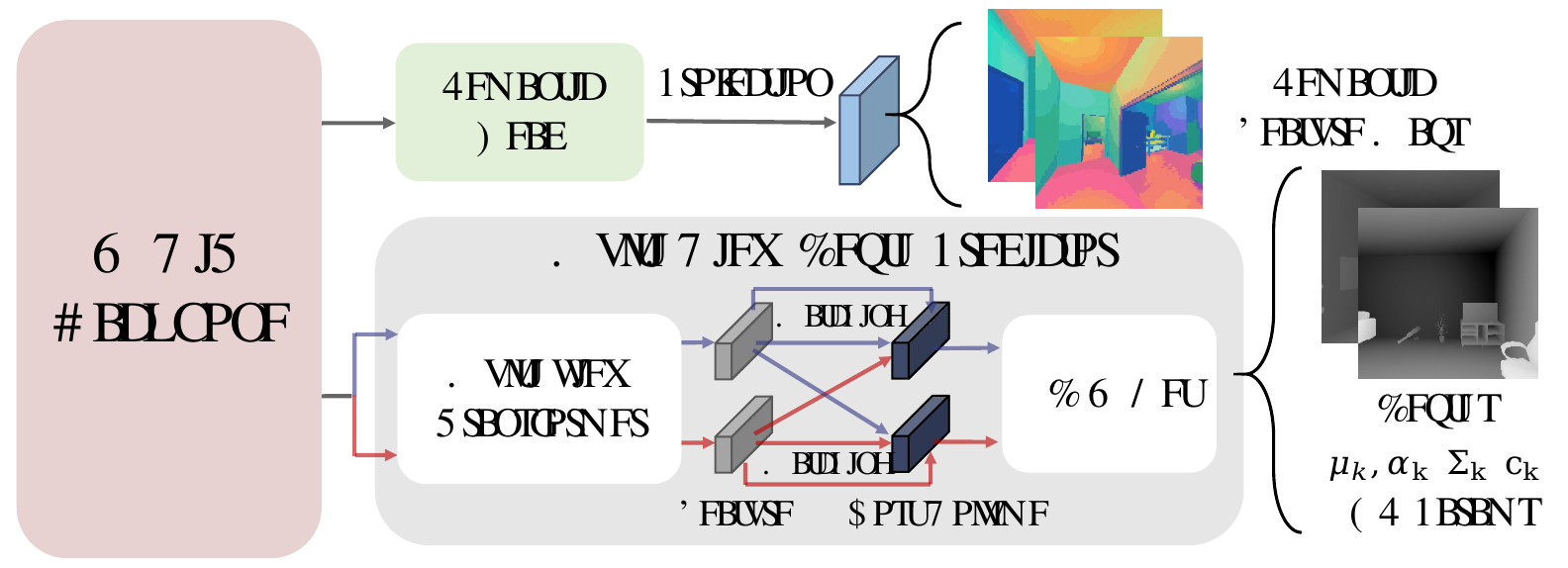}
    \vspace{-5pt}
    \caption{3D-Belief model architecture: 3D Scene Diffusion.}
    \label{fig:architecture_bottom}
    \vspace{-10pt}
\end{figure}

Our model (Figure~\ref{fig:architecture_bottom}) uses a shared U-ViT backbone \cite{song2025history, hoogeboom2024simpler} with two heads that jointly predict a 3DGS scene with semantics. For geometric consistency, we employ an MVS-style 3DGS predictor with a multi-view Transformer and cost-volume module, in which the cost volume stores cross-view feature matching scores over discretized depth candidates to guide depth prediction, producing depths and Gaussian parameters that are lifted into primitives for fast differentiable rendering \cite{chen2024mvsplat}. A lightweight semantic head linearly projects backbone features into per-pixel semantic maps, trained by distillation from CLIP-style embeddings \cite{kerr2023lerf}, enabling text-based querying at test time.

\subsection{Diffusion Training} \label{sec:diffusion_training}

Unlike generative models that perform \emph{frame-by-frame} diffusion in pixel or latent space, we adopt \emph{scene-level 3D diffusion} as introduced in~\cite{tewari2023diffusion}, applying diffusion to the entire 3D scene, i.e., the Gaussian primitives $z = {g_k}_{k}^K$ in this case. This design encourages global geometry modeling and multi-view consistency, yielding more effective 3D scene predictions for embodied tasks.

\(z\) is estimated by adding supervision with paired context images \(o^{\text{ctxt}}\) and target images \(o^{\text{trgt}}\). The forward process at diffusion time step $\tau$ is defined as:
\begin{equation}
    q(o^{\text{trgt}}_\tau \mid o^{\text{trgt}}_{\tau-1}) = \mathcal{N}(o^{\text{trgt}}_\tau ; \sqrt{1 - \beta_\tau} o^{\text{trgt}}_{\tau-1}, \beta_\tau I).
\end{equation}
In the reverse process, we reconstruct $o^{\text{trgt}}$ conditioned on $o^{\text{ctxt}}$ and target camera parameters $\phi^{\text{trgt}}$ by predicting a denoised scene $z_{\tau}$ and rendering it into observation space via the GS renderer \(\mathcal{G}(\cdot)\):
\begin{align}
    z_{\tau-1} &= \Phi_\theta(o^{\text{ctxt}}, o^{\text{trgt}}_\tau; \tau, \phi^{\text{ctxt}}, \phi^{\text{trgt}}), \label{eq:sample_z}\\
    \hat{o}^{\text{trgt}}_{\tau-1} &= \mathcal{G}(z_{\tau-1}, \phi^{\text{trgt}}).
    \label{eq:sample_o_hat}
\end{align}
Here, $\hat{o}^{\text{trgt}}_{t-1}$ serves as an estimate of the clean observation. $\Phi_\theta$ is the neural network predicting 3D Gaussians, as defined in Sec~\ref{sec:architecture}. See Appendix~\ref{appendix:diffusion} for the complete formulation.

\subsection{Training Objective} \label{sec:training_objective}
We train $\Phi_\theta$ with paired context and target observations.
Let $\hat{o}^{\text{trgt}}=\mathcal{G}(z_0,\phi^{\text{trgt}})$ and $\hat{o}^{\text{ctxt}}=\mathcal{G}(z_0,\phi^{\text{ctxt}})$ be renderings of the predicted scene in the target and context cameras.
For brevity, we use a view index $v\in\{\text{trgt},\text{ctxt}\}$ and let
$o^{v}\in\{o^{\text{trgt}},o^{\text{ctxt}}\}$ and $\hat{o}^{v}\in\{\hat{o}^{\text{trgt}},\hat{o}^{\text{ctxt}}\}$ denote the corresponding observation and rendering in view $v$. We use $I(\cdot)$ to denote RGB channels.

\textbf{RGB loss.} The RGB loss is
$\mathcal{L}_{\text{rgb}}
=
\sum_{v\in\{\text{trgt},\text{ctxt}\}}
\left\| I(\hat{o}^{v})-I(o^{v})\right\|_2^2$. 

\textbf{Semantic loss.}
We align the rendered semantic feature map with features extracted from image patches.
Let $S(\hat{o}^{v})\in\mathbb{R}^{H\times W\times d}$ be a per-pixel semantic feature map rendered from the predicted scene in view $v$.
For each $v$, we sample patch centers $\mathcal{P}^{v}=\{\mathbf{u}_j\}_{j=1}^M$ on $I(o^{v})$, crop patches $\pi(o^{v},\mathbf{u}_j)$,
and compute features using a frozen CLIP \cite{ilharco2021openclip} image encoder $f_{\text{clip}}(\cdot)$.
We supervise the semantic feature at the corresponding pixel using
$\mathcal{L}_{\text{sem}}
=
\sum_{v\in\{\text{trgt},\text{ctxt}\}}
\frac{1}{M}\sum_{j=1}^{M}
\left\|
S(\hat{o}^{v})(\mathbf{u}_j)
-
f_{\text{clip}}\!\left(\pi(o^{v},\mathbf{u}_j)\right)
\right\|_2^2$.

\textbf{Depth loss (optional).}
When ground-truth depth is available, we additionally supervise the rendered depth.
Let $D(\cdot)\in\mathbb{R}^{H\times W}$ denote the depth channel of an observation.
For each view $v$, define $\hat{d}^{v}=D(\hat{o}^{v})$ and $d^{v}=D(o^{v})$.
To handle missing or invalid depth values, let $m^{v}\in\{0,1\}^{H\times W}$ be a validity mask, where 1 indicates valid depth.
We use a masked $\ell_1$ loss:
$\mathcal{L}_{\text{depth}}
=
\sum_{v\in\{\text{trgt},\text{ctxt}\}}
\frac{1}{\sum_{\mathbf{u}} m^{v}(\mathbf{u})}
\sum_{\mathbf{u}}
m^{v}(\mathbf{u})
\left|
\hat{d}^{v}(\mathbf{u}) - d^{v}(\mathbf{u})
\right|$.

\section{Experiments}
\label{experiments}

We conduct a comprehensive evaluation of 3D-Belief as a generative 3D world model for embodied belief inference under partial observability. Our experiments evaluate its capabilities across 2D visual quality of belief prediction, 3D object and scene imagination, and belief-guided planning with inference-time scaling. Appendix~\ref{appendix:implementation} provides more implementation details.

\subsection{Experiment 1: 2D Visual Quality of Belief Prediction} \label{exp:vision}

Embodied belief requires both faithful memorization of observed regions and plausible imagination of unobserved space. One way to evaluate these capacities is to examine the visual quality of the belief predictions, similar to the common evaluations for video generation models \cite{ren2025gen3c}. A successful model needs to provide visual rendering of imagination that not only have rich visual details but also preserves the geometric and semantic integrity of the 3D scene.

\begin{figure*}[t!]
    \centering
    \includegraphics[width=1\linewidth]{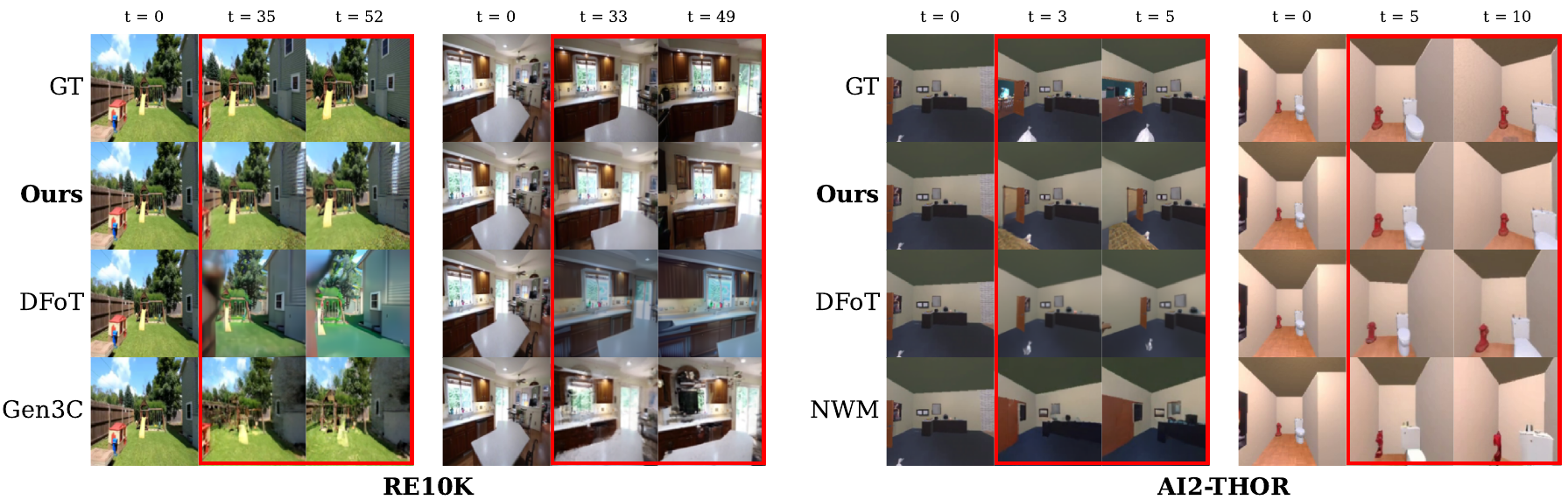}
    \vspace{-10pt}
    \caption{\textbf{Qualitative results of 2D visual quality of belief prediction.} Model predictions are highlighted with red boxes.
    We compare 3D-Belief against baselines on the 2D visual quality of belief prediction across AI2-THOR and RealEstate10K. 3D-Belief better preserves pose alignment, geometry, and texture details in both embodied and real-world trajectories.}
    \label{fig:2d_visual_belief_qual}
    \vspace{-10pt}
\end{figure*}

\textbf{Evaluation protocol.}
We use a two-part evaluation protocol. For observed regions, we condition the model on both ends of a trajectory and predict the intermediate novel views. Since these target views have paired ground truth, we evaluate them using LPIPS~\cite{zhang2018lpips}, PSNR, and SSIM. For unobserved regions, we condition the model only on the starting frame and generate views beyond the observed field of view. Because these predictions are inherently multimodal, we report distributional metrics, FVD~\cite{unterthiner2019fvd} and FID~\cite{heusel2017fid}, to measure the realism and consistency of the generated visual distribution. We evaluate on two domains. First, we use navigation datasets generated in AI2-THOR~\cite{kolve2017ai2} following~\cite{ehsani2024spoc}, which directly test embodied belief prediction under egocentric exploration and partial observability. On AI2-THOR, we compare against DFoT~\cite{song2025history} and NWM~\cite{bar2025navigation} fine-tuned on the training set. Second, we evaluate on RealEstate10K~\cite{zhou2018stereo}, a real-world visual prediction benchmark. This setting emphasizes photorealistic novel-view synthesis and video-level visual fidelity, complementing the embodied evaluation in AI2-THOR. On RealEstate10K, we compare against pretrained DFoT~\cite{song2025history} and GEN3C~\cite{ren2025gen3c}. Qualitative results from both datasets are shown in Figure~\ref{fig:2d_visual_belief_qual}.

\begin{table*}[t!]
\centering
\small
\caption{\textbf{2D visual quality of belief prediction across synthetic embodied and real-world domains.}
Scene memory is evaluated with paired metrics, while scene imagination is evaluated with distributional metrics. Best results within each domain are highlighted in \textbf{bold}.}
\vspace{10pt}
\setlength{\tabcolsep}{4pt}

\begin{minipage}{0.55\linewidth}
\centering
\begin{tabular}{llccc}
\toprule
\textbf{Domain} &
\textbf{Method} &
\textbf{PSNR}$\uparrow$ &
\textbf{SSIM}$\uparrow$ &
\textbf{LPIPS}$\downarrow$ \\
\midrule
\multirow{3}{*}{AI2-THOR}
& NWM  & 18.75 & 0.702 & 0.1876 \\
& DFoT & 23.35 & 0.841 & 0.1206 \\
& \textbf{Ours} & \textbf{28.81} & \textbf{0.928} & \textbf{0.0502} \\
\midrule
\multirow{3}{*}{RealEstate10K}
& DFoT  & 22.395 & 0.7167 & 0.1213 \\
& GEN3C & 23.104 & 0.7626 & 0.0978 \\
& \textbf{Ours} & \textbf{24.878} & \textbf{0.8432} & \textbf{0.0663} \\
\bottomrule
\end{tabular}
\subcaption{Observed scene}
\label{tab:2d_visual_observed}
\end{minipage}
\hspace{0.01\linewidth}
\begin{minipage}{0.40\linewidth}
\centering
\begin{tabular}{llcc}
\toprule
\textbf{Domain} &
\textbf{Method} &
\textbf{FID}$\downarrow$ &
\textbf{FVD}$\downarrow$ \\
\midrule
\multirow{3}{*}{AI2-THOR}
& NWM  & 89.28 & 487.4 \\
& DFoT & 72.82 & 429.7 \\
& \textbf{Ours} & \textbf{47.24} & \textbf{271.8} \\
\midrule
\multirow{3}{*}{RealEstate10K}
& DFoT  & 37.447 & 127.017 \\
& GEN3C & \textbf{24.291} & 73.744 \\
& \textbf{Ours} & 26.465 & \textbf{50.993} \\
\bottomrule
\end{tabular}
\subcaption{Imagined scene}
\label{tab:2d_visual_imagined}
\end{minipage}

\label{tab:2d_visual_belief}
\vspace{-20pt}
\end{table*}

\textbf{Results.}
Table~\ref{tab:2d_visual_belief} shows that 3D-Belief performs strongly across both embodied synthetic environments and real-world video trajectories. On AI2-THOR, our method consistently outperforms the baselines on scene memory and scene imagination, achieving the best paired reconstruction metrics and the best distributional metrics for imagined views. This indicates that explicit 3D belief modeling improves both faithful scene memory and plausible prediction beyond the observed field of view. On RealEstate10K, 3D-Belief further achieves the best PSNR, SSIM, LPIPS, and FVD among pretrained real-world visual prediction baselines, with competitive FID. These results show that 3D-Belief is not only effective in embodied partial-observation settings, but also generalizes to real-world camera trajectories, providing strong visual prediction quality across both domains.

Figure~\ref{fig:2d_visual_belief_qual} highlights two common baseline failure modes. First, baselines often exhibit \textit{semantic drift} beyond observed views. DFoT turns grass into a solid green ground in RealEstate10K, while AI2-THOR baselines predict opened doors without preserving plausible opening structure. In contrast, 3D-Belief better maintains semantic and scene consistency. Second, baselines suffer from \textit{pose and geometric drift} under viewpoint changes, causing generated content to shift or distort relative to the target view. This is especially visible in the right-hand AI2-THOR example, where object shapes and room layout degrade over time. 3D-Belief better preserves pose alignment and rigid scene structure through its explicit 3D belief representation.

\subsection{Experiment 2: 3D Object and Scene Imagination}\label{sec:reasoning}

\subsubsection{3D-CORE Benchmark}
In embodied tasks, agents must integrate partial observations over time, infer unobserved regions, and maintain stable beliefs while moving. For instance, object search requires reasoning about targets hidden behind occluders, while navigation benefits from inferring plausible free space and room layout from limited views. However, common evaluations of generative world models often focus on visual fidelity or short-horizon prediction~\cite{song2025history, bar2025navigation}, as in Sec.~\ref{exp:vision}. These metrics do not directly test whether a model forms consistent 3D beliefs under partial observability. We therefore introduce 3D-CORE (3D COntextual REasoning), a benchmark for embodied belief reasoning in 3D space. 3D-CORE tests: (1) \textbf{spatial expansion} beyond observed regions, (2) \textbf{semantic reasoning} grounded in 3D structure, and (3) \textbf{long-horizon consistency} under large viewpoint changes.

\textbf{Benchmark Construction.}
All Scenarios are built in AI2-THOR~\cite{kolve2017ai2} with ProcTHOR~\cite{NEURIPS2022_27c546ab} houses with diverse layouts and rich object assets from Objaverse~\cite{deitke2023objaverse}. We render egocentric RGB streams with known camera poses and provide ground-truth 3D geometry and semantics for evaluation.

\textbf{Tasks and Metrics.}
3D-CORE contains three complementary tasks that isolate distinct but essential aspects of contextual 3D reasoning. (See Appendix~\ref{appendix:metrics} for definitions of all metrics.)

\underline{Task 1: Object Completion (233 tasks):} The model observes a partially visible target object and completes its 3D geometry and appearance, producing plausible shape/texture consistent with the object category. Metrics: BEV IoU, 3D IoU, Chamfer distance, SigLIP similarity, and Recognition.

\underline{Task 2: Room Completion (263 tasks):} Given a single egocentric view of a room (with pose), the model predicts the unseen layout and semantics to support planning (e.g., plausible navigable regions). Metrics: Object Prediction F1, occupancy accuracy (known cells), and occupancy IoU.

\underline{Task 3: Object Permanence (474 tasks):} The model is rolled out along a trajectory with large viewpoint changes that returns to the start. We test whether the 3D belief stays stable, meaning that objects should not drift, or change identity upon revisiting. Metrics: LPIPS and SigLIP similarity.

\subsubsection{Evaluation on 3D-CORE}
We evaluate 3D-Belief's ability to imagine object- and scene-level 3D structure on the 3D-CORE benchmark. Given an initial RGB observation and an imagination camera trajectory, the agent predicts a completed 3D belief conditioned on the observation. Metrics are computed using semantics, geometry, or rendered views extracted from the predicted 3D representation.

\begin{figure}[t!]
  \centering
    \includegraphics[width=1.0\linewidth]{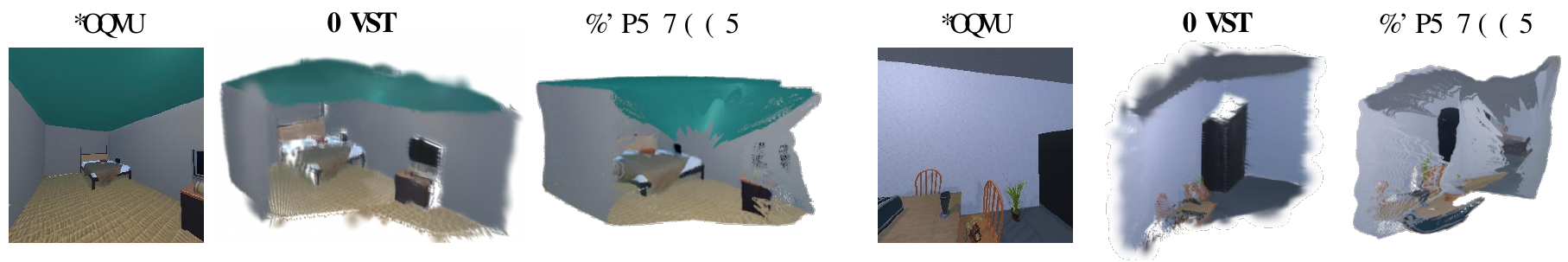}
    \vspace{-20pt}
    \caption{\textbf{Qualitative results of 3D object imagination.} Compared with the tested baseline, 3D-Belief better infers object-level semantic and geometric structure from the observed image, enabling more complete and consistent 3D imagination.}
  \label{fig:3d-core}
  \vspace{-10pt}
\end{figure}

\textbf{Baselines.} To the best of our knowledge, no existing method can directly imagine an explicit 3D scene. Thus, we construct a strong baseline, DFoT-VGGT, by combining two SOTA models. DFoT-VGGT is a ``imagination-then-lift'' baseline that first uses DFoT \cite{song2025history} to generate imagined observations and then applies VGGT \cite{wang2025vggt} to lift these predictions into a 3D representation.

\begin{table*}[t!]
\centering
\caption{Results on 3D-CORE: Object Completion, Room Completion, and Object Permanence.} \label{tab:3dcore}
\vspace{-5pt}

\resizebox{\textwidth}{!}{%
\begin{tabular}{lccccc ccc cc}
\toprule
& \multicolumn{5}{c}{\textbf{Object Completion (55\% Visibility)}}
& \multicolumn{3}{c}{\textbf{Room Completion}}
& \multicolumn{2}{c}{\textbf{Object Permanence}} \\
\cmidrule(lr){2-6}\cmidrule(lr){7-9}\cmidrule(lr){10-11}
\textbf{Method}
& \textbf{BEV IoU} $\uparrow$ & \textbf{3D IoU}$\uparrow$ & \textbf{Chamfer}$\downarrow$ & \textbf{SigLIP}$\uparrow$ & \textbf{Recognition}$\uparrow$
& \textbf{Obj. F1 $\uparrow$} & \textbf{Occ. Acc. $\uparrow$} & \textbf{Occ. IoU $\uparrow$}
& \textbf{LPIPS $\downarrow$} & \textbf{SigLIP $\uparrow$} \\
\midrule
DFoT-VGGT
& 0.362 & 0.243 & 0.830 & 0.798 & 0.767
& 0.531 & 0.252 & 0.110
& 0.555 & 0.907 \\
\textbf{3D Belief}
& \textbf{0.484} & \textbf{0.318} & \textbf{0.216} & \textbf{0.855} & \textbf{0.930}
& \textbf{0.536} & \textbf{0.900} & \textbf{0.442}
& \textbf{0.123} & \textbf{0.978} \\
\bottomrule
\end{tabular}%
}

\label{tab:combined-three-tasks}
\vspace{-10pt}
\end{table*}

\textbf{Results.} Table~\ref{tab:3dcore} summarizes the main results. Overall, 3D-Belief consistently outperforms DFoT-VGGT across all three 3D-CORE tasks, demonstrating stronger 3D belief reasoning for object completion, room-level spatial inference, and long-horizon consistency.

On \textbf{Object Completion} (full results in Appendix~\ref{appendix:objcomp_results}), 3D-Belief improves both geometry and semantics across visibility levels, with higher BEV/3D IoU, lower Chamfer distance, and better SigLIP similarity and VLM recognition. These gains indicate more faithful 3D completion and stronger preservation of object appearance and category semantics.

On \textbf{Room Completion}, the two models achieve similar object prediction F1, but 3D-Belief produces substantially more accurate occupancy beliefs, improving known-cell accuracy, free/occupied IoU, and overall occupancy IoU. This suggests stronger spatial belief estimation for downstream tasks.

On \textbf{Object Permanence}, 3D-Belief shows better long-horizon consistency under large camera motions, achieving higher perceptual and semantic consistency when returning to the initial viewpoint. This indicates reduced drift and more stable geometry and object identity over extended rollouts.

\textbf{Results.} Table~\ref{tab:3dcore} summarizes the quantitative results, and Figure~\ref{fig:3d-core} shows qualitative examples of object-level 3D imagination. Overall, 3D-Belief consistently outperforms DFoT-VGGT across all three 3D-CORE tasks, demonstrating stronger 3D belief reasoning for object completion, room-level spatial inference, and long-horizon consistency.

\subsection{Experiment 3: Belief-Guided Planning with Inference-Time Scaling}
\label{sec:exp-planning}

\begin{figure}[t!]
  \centering

  \begin{minipage}[t]{0.49\textwidth}
    \centering
    \includegraphics[width=\linewidth]{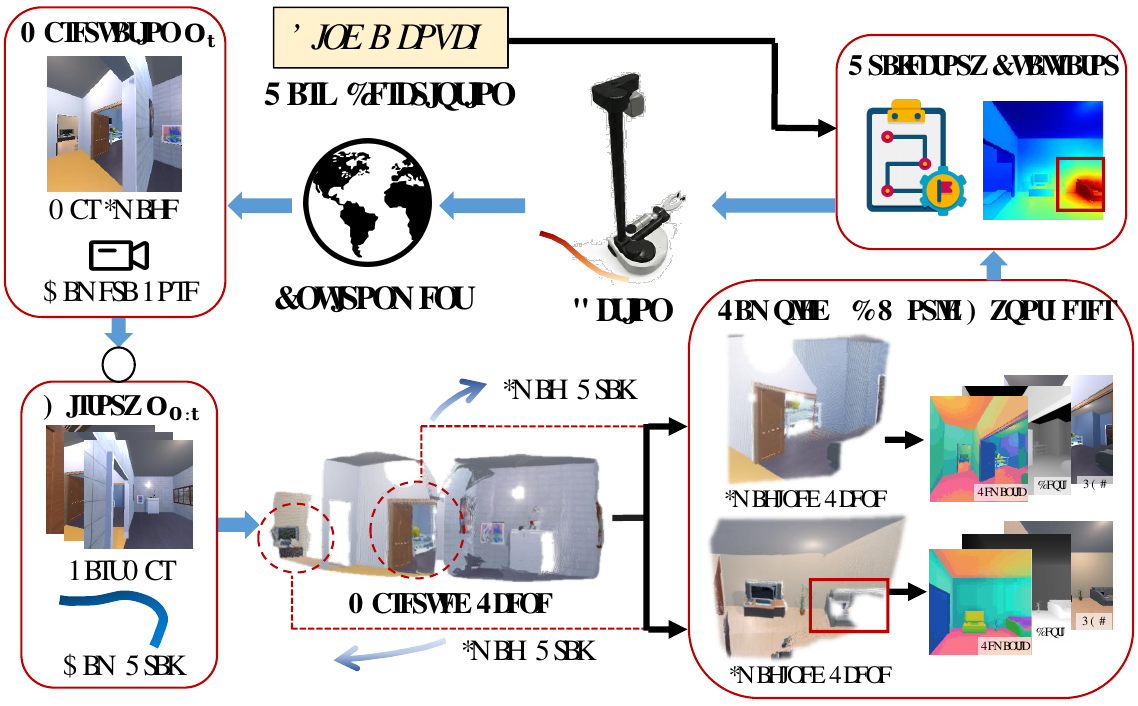}
    \captionof{figure}{\textbf{Belief-guided planning with 3D-Belief}. Dotted circles are imagination regions for visualization, and the red boxes are the potential target objects in imagined renders.}
    \label{fig:planning}
  \end{minipage}
  \hfill
  \begin{minipage}[t]{0.49\textwidth}
    \centering
    \includegraphics[width=\linewidth]{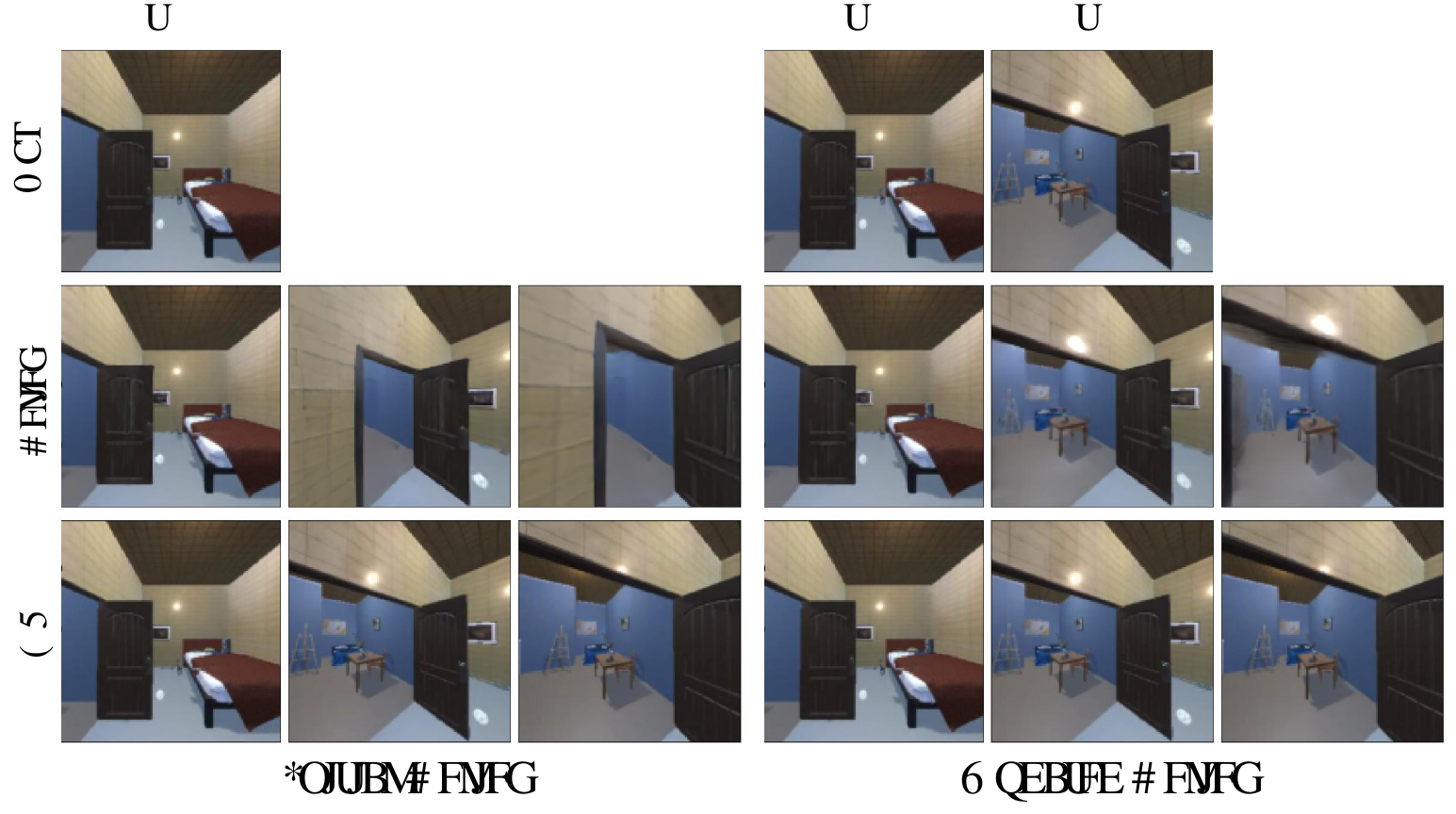}
    \captionof{figure}{\textbf{Belief updating during navigation}. The initial belief is formed from the first observation, while the updated belief incorporates later views, revising imagined regions and supporting navigation decisions.}
    \label{fig:sequential}
  \end{minipage}

  \vspace{-15pt}
\end{figure}

Embodied planning under partial observability requires an agent to act on both observed evidence and plausible unseen scene structure. We evaluate whether 3D-Belief can serve as a mental 3D world model for navigation. As shown in Figure~\ref{fig:planning}, at each step, the agent updates its belief from egocentric RGB history and camera poses, samples multiple plausible 3D beliefs, and evaluates candidate paths by rendering imagined observations along them. Rollouts are scored with open-vocabulary semantic queries on the imagined belief, guiding the planner toward paths likely to reveal or approach the target. Figure~\ref{fig:sequential} illustrates how 3D-Belief updates the agent's belief during navigation. Additional implementation details are provided in Appendix~\ref{appendix:planning}.

For simulation, we train 3D-Belief on AI2-THOR~\cite{kolve2017ai2} navigation episodes; for real-world experiments, we train a separate model on RealEstate10K~\cite{zhou2018stereo}. To support open-vocabulary goals, semantic heads are trained with OpenCLIP~\cite{ilharco2021openclip} features and regularized with DINOv3~\cite{simeoni2025dinov3}, following LERF~\cite{kerr2023lerf}. This enables \textit{inference-time scaling}: the agent samples multiple beliefs, evaluates multiple imagined paths, and selects the rollout with the strongest goal progress under uncertainty.

\subsubsection{Object Navigation in Simulations} \label{exp:planning_sim}

\textbf{Simulation Environment.}
We conduct all simulated tasks following \cite{ehsani2024spoc} in the AI2-THOR simulator~\cite{kolve2017ai2}, using ProcTHOR house assets~\cite{NEURIPS2022_27c546ab}.
Unless otherwise noted, we evaluate on 135 unseen houses for test, ensuring generalization to novel layouts and object configurations.

\textbf{Task Definition.}
We define the task following \cite{ehsani2024spoc}. Each episode starts the agent at a random location with a target object category, which may require multi-room exploration. The agent receives egocentric RGB observations and poses sequentially and must plan actions online. Success requires the agent to be within a proximity threshold of the target and to see it in the central region of view.

\textbf{Metrics.}
We evaluate task performance using Success Rate (SR) for overall success, Success-weighted by Path Length (SPL) for path efficiency, Success-weighted by Episode Length (SEL) \cite{eftekhar2023selective,yokoyama2021success} for time efficiency. We also report token costs for VLM usage.

\begin{table*}[t]
\centering
\caption{\textbf{Object navigation results in simulations and the real world.} More results are provided in Appendix~\ref{appendix:planning_results}.}
\vspace{-5pt}
\label{tab:objsearch-results}
\scriptsize
\setlength{\tabcolsep}{3pt}
\renewcommand{\arraystretch}{1.15}

\resizebox{\linewidth}{!}{
\begin{tabular}{lcccccccc|ccc}
\toprule
& \multicolumn{8}{c|}{\textbf{Simulations}}
& \multicolumn{3}{c}{\textbf{Real World}} \\
\cmidrule(lr){2-9}\cmidrule(lr){10-12}
\textbf{Metric} &
\makecell{VGGT\\(w/ frontier)} &
\makecell{VGGT\\(w/ Gemini 3.0)} &
\makecell{DFoT-VGGT\\(w/ Gemini 3.0)} &
\makecell{NWM-VGGT\\(w/ Gemini 3.0)} &
\makecell{Gemini 3.0} &
\makecell{Qwen3-VL-8B\\Instruct} &
\makecell{SPOC} &
\makecell{\textbf{Ours}} &
\makecell{Gemini 3.0} &
\makecell{\textbf{Ours}} &
\makecell{} \\
\midrule
SR\% $\uparrow$
& 27.50 & 25.00 & 26.05 & 25.00 & 45.00 & 18.33 & 31.67 & \textbf{59.17}
& 23.08 & \textbf{55.56} & \\

SPL\% $\uparrow$
& 25.82 & 24.10 & 24.59 & 23.46 & 37.81 & 14.08 & 30.97 & \textbf{39.07}
& -- & -- & \\

SEL\% $\uparrow$
& 22.66 & 22.66 & 23.43 & 20.76 & \textbf{41.47} & 16.94 & 30.56 & 40.24
& 13.55 & \textbf{35.91} & \\

Token (/step) $\downarrow$
& \textbf{0} & 1448.02 & 1210.08 & 552.12 & 7512.70 & 220.29 & \textbf{0} & \textbf{0}
& 2317.09 & \textbf{0} & \\
\bottomrule
\end{tabular}
}
\vspace{-10pt}
\end{table*}

\textbf{Baselines.}
We compare 3D-Belief against four categories of methods, as shown below:

\begin{addmargin}[-25pt]{0pt} 
\begin{itemize}\setlength\itemsep{0pt}

\item \underline{3D Reconstruction.}
This group tests whether high-quality \textbf{reconstruction of only the observed geometry} is sufficient for efficient search.
We use VGGT~\cite{wang2025vggt} to build a 3D cache from the agent’s RGB history.
Variants differ only in the goal-selection module: a classical frontier-based explorer (VGGT w/ frontier) and VLM-driven waypoint selection.

\item \underline{Lifted 2D Imagination.}
This group tests whether imagined future observations can be lifted into 3D for downstream navigation.
We first use video prediction models, DFoT and NWM, to imagine future RGB observations, and then use VGGT to reconstruct a 3D representation from the generated views.
Goal selection is performed by a VLM, either GPT-5 mini or Gemini 3.0.

\item \underline{VLM Agents.}
This group evaluates whether end-to-end VLM-based agents can directly solve object navigation from visual observations.
We compare against GPT-5 mini, Gemini 3.0, and Qwen3-VL-8B-Instruct, which select actions or subgoals based on the agent’s observation history.

\item \underline{Policy Model.}
This group evaluates a learned navigation policy trained for embodied object search.
We include SPOC~\cite{ehsani2024spoc}, which provides a strong policy-based baseline without explicit VLM reasoning during online inference.

\end{itemize}
\end{addmargin}

\textbf{Results.} Results in Table~\ref{tab:objsearch-results} show that 3D-Belief consistently outperforms all baseline categories, highlighting the benefit of an explicit, updatable 3D belief for planning. It achieves the best SR and strongest SPL/SEL, indicating higher success with more efficient paths and shorter completion times. 3D reconstruction baselines (VGGT variants) are limited by incomplete beliefs about unseen regions. Imagination-then-lift pipelines (DFoT-VGGT, NWM-VGGT) improve some efficiency metrics but suffer from decoupled stages and long-horizon inconsistency. VLM agents can succeed via semantic reasoning but are less stable and far more expensive, with worse SPL and high token costs.

\subsubsection{Object Navigation in the Real World} \label{exp:planning_real}

We evaluate 3D-Belief on a real mobile manipulation platform (Hello Robot Stretch) in a mock apartment environment.
The environment contains typical household furniture and objects, forming realistic object navigation scenarios. See Appendix~\ref{appendix:real} for an example environment setup. Note that the environments, objects, and target descriptions are all unseen during training, making this a real-world open-vocabulary object navigation setting.

\textbf{Success Criteria.}
At the beginning of each episode, the robot is given the name of the target object (e.g., red mug), and must explore the environment and stop when it is found.
An episode is marked as successful if (1) the robot reaches within a distance threshold of the target and (2) the target is within the central region of the egocentric view with a facing angle within a tolerance.

\textbf{Baseline.}
We test the best performing baseline in simulations, VLM agent based on Gemini 3.0.

\textbf{Result.} Results are shown in Table~\ref{tab:objsearch-results}. Across real-robot episodes, 3D-Belief enables more reliable and efficient performance than the Gemini 3.0 agent.
It achieves a higher success rate and improved efficiency, demonstrating that explicit, online-updatable 3D beliefs transfer to real-world settings and can robustly support embodied decision making under sensor noise.

\section{Conclusion}
\label{conclusion}
In this work, we proposed a new framework for embodied belief inference via generative 3D world modeling and identified key capabilities for practical 3D belief modeling. We then proposed 3D-Belief, which predicts unseen regions in an explicit 3D representation from partial observations and updates this belief online. Experimental results on 2D visual quality, contextual reasoning, and object navigation have demonstrated that 3D-Belief improves upon existing generative world models in 2D and 3D scene memory and imagination, as well as downstream embodied task performance.

\textbf{Limitations and Future Work.} Our 3D-Belief model assumes a static world. Future work will incorporate dynamic world modeling to support broader embodied tasks. Another direction is to improve the controllability of 3D imagination by conditioning hypothesis sampling on high-level guidance, such as language descriptions or scene graphs of the imagined 3D world.

\section*{Acknowledgment}
TS acknowledges the computing resources provided by NAIRR Pilot.

\bibliography{references}

@inproceedings{ehsani2024spoc,
  title={SPOC: Imitating Shortest Paths in Simulation Enables Effective Navigation and Manipulation in the Real World},
  author={Ehsani, Kiana and Gupta, Tanmay and Hendrix, Rose and Salvador, Jordi and Weihs, Luca and Zeng, Kuo-Hao and Singh, Kunal Pratap and Kim, Yejin and Han, Winson and Herrasti, Alvaro and others},
  booktitle={Proceedings of the IEEE/CVF Conference on Computer Vision and Pattern Recognition},
  pages={16238--16250},
  year={2024}
}

@inproceedings{ramakrishnan2021hm3d,
  title={Habitat-Matterport 3D Dataset ({HM}3D): 1000 Large-scale 3D Environments for Embodied {AI}},
  author={Santhosh Kumar Ramakrishnan and Aaron Gokaslan and Erik Wijmans and Oleksandr Maksymets and Alexander Clegg and John M Turner and Eric Undersander and Wojciech Galuba and Andrew Westbury and Angel X Chang and Manolis Savva and Yili Zhao and Dhruv Batra},
  booktitle={Thirty-fifth Conference on Neural Information Processing Systems Datasets and Benchmarks Track},
  year={2021}
}

@article{tewari2023diffusion,
  title={Diffusion with forward models: Solving stochastic inverse problems without direct supervision},
  author={Tewari, Ayush and Yin, Tianwei and Cazenavette, George and Rezchikov, Semon and Tenenbaum, Josh and Durand, Fr{\'e}do and Freeman, Bill and Sitzmann, Vincent},
  journal={Advances in Neural Information Processing Systems},
  volume={36},
  pages={12349--12362},
  year={2023}
}

@article{kaelbling1998planning,
  title={Planning and acting in partially observable stochastic domains},
  author={Kaelbling, Leslie Pack and Littman, Michael L and Cassandra, Anthony R},
  journal={Artificial intelligence},
  volume={101},
  number={1-2},
  pages={99--134},
  year={1998},
  publisher={Elsevier}
}

@article{zhou2024robodreamer,
  title={Robodreamer: Learning compositional world models for robot imagination},
  author={Zhou, Siyuan and Du, Yilun and Chen, Jiaben and Li, Yandong and Yeung, Dit-Yan and Gan, Chuang},
  journal={arXiv preprint arXiv:2404.12377},
  year={2024}
}

@article{zhang2024combo,
  title={COMBO: compositional world models for embodied multi-agent cooperation},
  author={Zhang, Hongxin and Wang, Zeyuan and Lyu, Qiushi and Zhang, Zheyuan and Chen, Sunli and Shu, Tianmin and Dariush, Behzad and Lee, Kwonjoon and Du, Yilun and Gan, Chuang},
  journal={arXiv preprint arXiv:2404.10775},
  year={2024}
}

@inproceedings{
du2024video,
title={Video Language Planning},
author={Yilun Du and Sherry Yang and Pete Florence and Fei Xia and Ayzaan Wahid and brian ichter and Pierre Sermanet and Tianhe Yu and Pieter Abbeel and Joshua B. Tenenbaum and Leslie Pack Kaelbling and Andy Zeng and Jonathan Tompson},
booktitle={The Twelfth International Conference on Learning Representations},
year={2024}
}

@article{zhen2025tesseract,
  title={TesserAct: Learning 4D Embodied World Models},
  author={Zhen, Haoyu and Sun, Qiao and Zhang, Hongxin and Li, Junyan and Zhou, Siyuan and Du, Yilun and Gan, Chuang},
  journal={arXiv preprint arXiv:2504.20995},
  year={2025}
}

@article{song2025history,
  title={History-Guided Video Diffusion},
  author={Song, Kiwhan and Chen, Boyuan and Simchowitz, Max and Du, Yilun and Tedrake, Russ and Sitzmann, Vincent},
  journal={arXiv preprint arXiv:2502.06764},
  year={2025}
}

@inproceedings{wang2025vggt,
  title={Vggt: Visual geometry grounded transformer},
  author={Wang, Jianyuan and Chen, Minghao and Karaev, Nikita and Vedaldi, Andrea and Rupprecht, Christian and Novotny, David},
  booktitle={Proceedings of the Computer Vision and Pattern Recognition Conference},
  pages={5294--5306},
  year={2025}
}

@inproceedings{yokoyama2021success,
  title={Success weighted by completion time: A dynamics-aware evaluation criteria for embodied navigation},
  author={Yokoyama, Naoki and Ha, Sehoon and Batra, Dhruv},
  booktitle={2021 IEEE/RSJ International Conference on Intelligent Robots and Systems (IROS)},
  pages={1562--1569},
  year={2021},
  organization={IEEE}
}

@article{eftekhar2023selective,
  title={Selective visual representations improve convergence and generalization for embodied ai},
  author={Eftekhar, Ainaz and Zeng, Kuo-Hao and Duan, Jiafei and Farhadi, Ali and Kembhavi, Ani and Krishna, Ranjay},
  journal={arXiv preprint arXiv:2311.04193},
  year={2023}
}

@inproceedings{bar2025navigation,
  title={Navigation world models},
  author={Bar, Amir and Zhou, Gaoyue and Tran, Danny and Darrell, Trevor and LeCun, Yann},
  booktitle={Proceedings of the Computer Vision and Pattern Recognition Conference},
  pages={15791--15801},
  year={2025}
}

@article{wu2025geometry,
  title={Geometry forcing: Marrying video diffusion and 3d representation for consistent world modeling},
  author={Wu, Haoyu and Wu, Diankun and He, Tianyu and Guo, Junliang and Ye, Yang and Duan, Yueqi and Bian, Jiang},
  journal={arXiv preprint arXiv:2507.07982},
  year={2025}
}

@article{hafner2025training,
  title={Training agents inside of scalable world models},
  author={Hafner, Danijar and Yan, Wilson and Lillicrap, Timothy},
  journal={arXiv preprint arXiv:2509.24527},
  year={2025}
}

@inproceedings{wang2024dust3r,
  title={Dust3r: Geometric 3d vision made easy},
  author={Wang, Shuzhe and Leroy, Vincent and Cabon, Yohann and Chidlovskii, Boris and Revaud, Jerome},
  booktitle={Proceedings of the IEEE/CVF Conference on Computer Vision and Pattern Recognition},
  pages={20697--20709},
  year={2024}
}

@article{zhuo2025streaming,
  title={Streaming 4d visual geometry transformer},
  author={Zhuo, Dong and Zheng, Wenzhao and Guo, Jiahe and Wu, Yuqi and Zhou, Jie and Lu, Jiwen},
  journal={arXiv preprint arXiv:2507.11539},
  year={2025}
}

@inproceedings{wang2025continuous,
  title={Continuous 3d perception model with persistent state},
  author={Wang, Qianqian and Zhang, Yifei and Holynski, Aleksander and Efros, Alexei A and Kanazawa, Angjoo},
  booktitle={Proceedings of the Computer Vision and Pattern Recognition Conference},
  pages={10510--10522},
  year={2025}
}

@inproceedings{kerr2023lerf,
  title={Lerf: Language embedded radiance fields},
  author={Kerr, Justin and Kim, Chung Min and Goldberg, Ken and Kanazawa, Angjoo and Tancik, Matthew},
  booktitle={Proceedings of the IEEE/CVF international conference on computer vision},
  pages={19729--19739},
  year={2023}
}

@inproceedings{gu2024conceptgraphs,
  title={Conceptgraphs: Open-vocabulary 3d scene graphs for perception and planning},
  author={Gu, Qiao and Kuwajerwala, Ali and Morin, Sacha and Jatavallabhula, Krishna Murthy and Sen, Bipasha and Agarwal, Aditya and Rivera, Corban and Paul, William and Ellis, Kirsty and Chellappa, Rama and others},
  booktitle={2024 IEEE International Conference on Robotics and Automation (ICRA)},
  pages={5021--5028},
  year={2024},
  organization={IEEE}
}

@inproceedings{liu2025dynamem,
  title={Dynamem: Online dynamic spatio-semantic memory for open world mobile manipulation},
  author={Liu, Peiqi and Guo, Zhanqiu and Warke, Mohit and Chintala, Soumith and Paxton, Chris and Shafiullah, Nur Muhammad Mahi and Pinto, Lerrel},
  booktitle={2025 IEEE International Conference on Robotics and Automation (ICRA)},
  pages={13346--13355},
  year={2025},
  organization={IEEE}
}

@article{kolve2017ai2,
  title={Ai2-thor: An interactive 3d environment for visual ai},
  author={Kolve, Eric and Mottaghi, Roozbeh and Han, Winson and VanderBilt, Eli and Weihs, Luca and Herrasti, Alvaro and Deitke, Matt and Ehsani, Kiana and Gordon, Daniel and Zhu, Yuke and others},
  journal={arXiv preprint arXiv:1712.05474},
  year={2017}
}

@inproceedings{ren2025gen3c,
  title={Gen3c: 3d-informed world-consistent video generation with precise camera control},
  author={Ren, Xuanchi and Shen, Tianchang and Huang, Jiahui and Ling, Huan and Lu, Yifan and Nimier-David, Merlin and M{\"u}ller, Thomas and Keller, Alexander and Fidler, Sanja and Gao, Jun},
  booktitle={Proceedings of the Computer Vision and Pattern Recognition Conference},
  pages={6121--6132},
  year={2025}
}

@article{yu2024viewcrafter,
  title={Viewcrafter: Taming video diffusion models for high-fidelity novel view synthesis},
  author={Yu, Wangbo and Xing, Jinbo and Yuan, Li and Hu, Wenbo and Li, Xiaoyu and Huang, Zhipeng and Gao, Xiangjun and Wong, Tien-Tsin and Shan, Ying and Tian, Yonghong},
  journal={arXiv preprint arXiv:2409.02048},
  year={2024}
}

@article{yu2025trajectorycrafter,
  title={Trajectorycrafter: Redirecting camera trajectory for monocular videos via diffusion models},
  author={Yu, Mark and Hu, Wenbo and Xing, Jinbo and Shan, Ying},
  journal={arXiv preprint arXiv:2503.05638},
  year={2025}
}

@inproceedings{charatan2024pixelsplat,
  title={pixelsplat: 3d gaussian splats from image pairs for scalable generalizable 3d reconstruction},
  author={Charatan, David and Li, Sizhe Lester and Tagliasacchi, Andrea and Sitzmann, Vincent},
  booktitle={Proceedings of the IEEE/CVF conference on computer vision and pattern recognition},
  pages={19457--19467},
  year={2024}
}

@inproceedings{chen2024mvsplat,
  title={Mvsplat: Efficient 3d gaussian splatting from sparse multi-view images},
  author={Chen, Yuedong and Xu, Haofei and Zheng, Chuanxia and Zhuang, Bohan and Pollefeys, Marc and Geiger, Andreas and Cham, Tat-Jen and Cai, Jianfei},
  booktitle={European Conference on Computer Vision},
  pages={370--386},
  year={2024},
  organization={Springer}
}

@article{ye2024no,
  title={No pose, no problem: Surprisingly simple 3d gaussian splats from sparse unposed images},
  author={Ye, Botao and Liu, Sifei and Xu, Haofei and Li, Xueting and Pollefeys, Marc and Yang, Ming-Hsuan and Peng, Songyou},
  journal={arXiv preprint arXiv:2410.24207},
  year={2024}
}

@article{jiang2025anysplat,
  title={Anysplat: Feed-forward 3d gaussian splatting from unconstrained views},
  author={Jiang, Lihan and Mao, Yucheng and Xu, Linning and Lu, Tao and Ren, Kerui and Jin, Yichen and Xu, Xudong and Yu, Mulin and Pang, Jiangmiao and Zhao, Feng and others},
  journal={ACM Transactions on Graphics (TOG)},
  volume={44},
  number={6},
  pages={1--16},
  year={2025},
  publisher={ACM New York, NY, USA}
}

@inproceedings{gu2025diffusion,
  title={Diffusion as shader: 3d-aware video diffusion for versatile video generation control},
  author={Gu, Zekai and Yan, Rui and Lu, Jiahao and Li, Peng and Dou, Zhiyang and Si, Chenyang and Dong, Zhen and Liu, Qifeng and Lin, Cheng and Liu, Ziwei and others},
  booktitle={Proceedings of the Special Interest Group on Computer Graphics and Interactive Techniques Conference Conference Papers},
  pages={1--12},
  year={2025}
}

@inproceedings{ma2025you,
  title={You see it, you got it: Learning 3d creation on pose-free videos at scale},
  author={Ma, Baorui and Gao, Huachen and Deng, Haoge and Luo, Zhengxiong and Huang, Tiejun and Tang, Lulu and Wang, Xinlong},
  booktitle={Proceedings of the Computer Vision and Pattern Recognition Conference},
  pages={2016--2029},
  year={2025}
}

@inproceedings{wu2025difix3d+,
  title={Difix3d+: Improving 3d reconstructions with single-step diffusion models},
  author={Wu, Jay Zhangjie and Zhang, Yuxuan and Turki, Haithem and Ren, Xuanchi and Gao, Jun and Shou, Mike Zheng and Fidler, Sanja and Gojcic, Zan and Ling, Huan},
  booktitle={Proceedings of the Computer Vision and Pattern Recognition Conference},
  pages={26024--26035},
  year={2025}
}

@article{huang2025voyager,
  title={Voyager: Long-range and world-consistent video diffusion for explorable 3d scene generation},
  author={Huang, Tianyu and Zheng, Wangguandong and Wang, Tengfei and Liu, Yuhao and Wang, Zhenwei and Wu, Junta and Jiang, Jie and Li, Hui and Lau, Rynson and Zuo, Wangmeng and others},
  journal={ACM Transactions on Graphics (TOG)},
  volume={44},
  number={6},
  pages={1--15},
  year={2025},
  publisher={ACM New York, NY, USA}
}

@inproceedings{cao2025uni3c,
  title={Uni3c: Unifying precisely 3d-enhanced camera and human motion controls for video generation},
  author={Cao, Chenjie and Zhou, Jingkai and Li, Shikai and Liang, Jingyun and Yu, Chaohui and Wang, Fan and Xue, Xiangyang and Fu, Yanwei},
  booktitle={Proceedings of the SIGGRAPH Asia 2025 Conference Papers},
  pages={1--12},
  year={2025}
}

@article{kerbl20233d,
  title={3D Gaussian splatting for real-time radiance field rendering.},
  author={Kerbl, Bernhard and Kopanas, Georgios and Leimk{\"u}hler, Thomas and Drettakis, George},
  journal={ACM Trans. Graph.},
  volume={42},
  number={4},
  pages={139--1},
  year={2023}
}

@article{qiu2024feature,
  title={Feature splatting: Language-driven physics-based scene synthesis and editing},
  author={Qiu, Ri-Zhao and Yang, Ge and Zeng, Weijia and Wang, Xiaolong},
  journal={arXiv preprint arXiv:2404.01223},
  year={2024}
}

@article{zhou2018stereo,
  title={Stereo magnification: Learning view synthesis using multiplane images},
  author={Zhou, Tinghui and Tucker, Richard and Flynn, John and Fyffe, Graham and Snavely, Noah},
  journal={arXiv preprint arXiv:1805.09817},
  year={2018}
}

@inproceedings{ling2024dl3dv,
  title={Dl3dv-10k: A large-scale scene dataset for deep learning-based 3d vision},
  author={Ling, Lu and Sheng, Yichen and Tu, Zhi and Zhao, Wentian and Xin, Cheng and Wan, Kun and Yu, Lantao and Guo, Qianyu and Yu, Zixun and Lu, Yawen and others},
  booktitle={Proceedings of the IEEE/CVF Conference on Computer Vision and Pattern Recognition},
  pages={22160--22169},
  year={2024}
}

@inproceedings{hoogeboom2023simple,
  title={simple diffusion: End-to-end diffusion for high resolution images},
  author={Hoogeboom, Emiel and Heek, Jonathan and Salimans, Tim},
  booktitle={International Conference on Machine Learning},
  pages={13213--13232},
  year={2023},
  organization={PMLR}
}

@article{hoogeboom2024simpler,
  title={Simpler diffusion (sid2): 1.5 fid on imagenet512 with pixel-space diffusion},
  author={Hoogeboom, Emiel and Mensink, Thomas and Heek, Jonathan and Lamerigts, Kay and Gao, Ruiqi and Salimans, Tim},
  journal={arXiv preprint arXiv:2410.19324},
  year={2024}
}

@article{ilharco2021openclip,
  title={Openclip},
  author={Ilharco, Gabriel and Wortsman, Mitchell and Carlini, Nicholas and Taori, Rohan and Dave, Achal and Shankar, Vaishaal and Namkoong, Hongseok and Miller, John and Hajishirzi, Hannaneh and Farhadi, Ali and others},
  journal={Zenodo},
  year={2021}
}

@inproceedings{NEURIPS2022_27c546ab,
 author = {Deitke, Matt and VanderBilt, Eli and Herrasti, Alvaro and Weihs, Luca and Ehsani, Kiana and Salvador, Jordi and Han, Winson and Kolve, Eric and Kembhavi, Aniruddha and Mottaghi, Roozbeh},
 booktitle = {Advances in Neural Information Processing Systems},
 editor = {S. Koyejo and S. Mohamed and A. Agarwal and D. Belgrave and K. Cho and A. Oh},
 pages = {5982--5994},
 publisher = {Curran Associates, Inc.},
 title = {ProcTHOR: Large-Scale Embodied AI Using Procedural Generation},
 volume = {35},
 year = {2022}
}

@article{xiao2025worldmem,
  title={Worldmem: Long-term consistent world simulation with memory},
  author={Xiao, Zeqi and Lan, Yushi and Zhou, Yifan and Ouyang, Wenqi and Yang, Shuai and Zeng, Yanhong and Pan, Xingang},
  journal={arXiv preprint arXiv:2504.12369},
  year={2025}
}

@article{melnik2024video,
  title={Video diffusion models: A survey},
  author={Melnik, Andrew and Ljubljanac, Michal and Lu, Cong and Yan, Qi and Ren, Weiming and Ritter, Helge},
  journal={arXiv preprint arXiv:2405.03150},
  year={2024}
}

@article{oshima2025worldpack,
  title={WorldPack: Compressed Memory Improves Spatial Consistency in Video World Modeling},
  author={Oshima, Yuta and Iwasawa, Yusuke and Suzuki, Masahiro and Matsuo, Yutaka and Furuta, Hiroki},
  journal={arXiv preprint arXiv:2512.02473},
  year={2025}
}

@article{yang2025mindjourney,
  title={MindJourney: Test-Time Scaling with World Models for Spatial Reasoning},
  author={Yang, Yuncong and Liu, Jiageng and Zhang, Zheyuan and Zhou, Siyuan and Tan, Reuben and Yang, Jianwei and Du, Yilun and Gan, Chuang},
  journal={arXiv preprint arXiv:2507.12508},
  year={2025}
}

@article{zhou2025stable,
  title={Stable virtual camera: Generative view synthesis with diffusion models},
  author={Zhou, Jensen and Gao, Hang and Voleti, Vikram and Vasishta, Aaryaman and Yao, Chun-Han and Boss, Mark and Torr, Philip and Rupprecht, Christian and Jampani, Varun},
  journal={arXiv preprint arXiv:2503.14489},
  year={2025}
}

@article{du2023video,
  title={Video Language Planning},
  author={Du, Yilun and Yang, Mengjiao and Florence, Pete and Xia, Fei and Wahid, Ayzaan and Ichter, Brian and Sermanet, Pierre and Yu, Tianhe and Abbeel, Pieter and Tenenbaum, Joshua B and others},
  journal={arXiv preprint arXiv:2310.10625},
  year={2023}
}

@article{karnan2022scand,
  title={Socially compliant navigation dataset (scand): A large-scale dataset of demonstrations for social navigation},
  author={Karnan, Haresh and Nair, Anirudh and Xiao, Xuesu and Warnell, Garrett and Pirk, S{\"o}ren and Toshev, Alexander and Hart, Justin and Biswas, Joydeep and Stone, Peter},
  journal={IEEE Robotics and Automation Letters},
  volume={7},
  number={4},
  pages={11807--11814},
  year={2022},
  publisher={IEEE}
}

@inproceedings{triest2022tartandrive,
  title={Tartandrive: A large-scale dataset for learning off-road dynamics models},
  author={Triest, Samuel and Sivaprakasam, Matthew and Wang, Sean J and Wang, Wenshan and Johnson, Aaron M and Scherer, Sebastian},
  booktitle={2022 International Conference on Robotics and Automation (ICRA)},
  pages={2546--2552},
  year={2022},
  organization={IEEE}
}

@article{shah2021recon,
  title={Rapid exploration for open-world navigation with latent goal models},
  author={Shah, Dhruv and Eysenbach, Benjamin and Kahn, Gregory and Rhinehart, Nicholas and Levine, Sergey},
  journal={arXiv preprint arXiv:2104.05859},
  year={2021}
}

@article{hirose2023sacson,
  title={Sacson: Scalable autonomous control for social navigation},
  author={Hirose, Noriaki and Shah, Dhruv and Sridhar, Ajay and Levine, Sergey},
  journal={IEEE Robotics and Automation Letters},
  volume={9},
  number={1},
  pages={49--56},
  year={2023},
  publisher={IEEE}
}

@inproceedings{grauman2022ego4d,
  title={Ego4d: Around the world in 3,000 hours of egocentric video},
  author={Grauman, Kristen and Westbury, Andrew and Byrne, Eugene and Chavis, Zachary and Furnari, Antonino and Girdhar, Rohit and Hamburger, Jackson and Jiang, Hao and Liu, Miao and Liu, Xingyu and others},
  booktitle={Proceedings of the IEEE/CVF conference on computer vision and pattern recognition},
  pages={18995--19012},
  year={2022}
}

@inproceedings{zhang2018lpips,
  title={The unreasonable effectiveness of deep features as a perceptual metric},
  author={Zhang, Richard and Isola, Phillip and Efros, Alexei A and Shechtman, Eli and Wang, Oliver},
  booktitle={Proceedings of the IEEE conference on computer vision and pattern recognition},
  pages={586--595},
  year={2018}
}

@article{heusel2017fid,
  title={Gans trained by a two time-scale update rule converge to a local nash equilibrium},
  author={Heusel, Martin and Ramsauer, Hubert and Unterthiner, Thomas and Nessler, Bernhard and Hochreiter, Sepp},
  journal={Advances in neural information processing systems},
  volume={30},
  year={2017}
}

@article{unterthiner2019fvd,
  title={FVD: A new metric for video generation},
  author={Unterthiner, Thomas and Van Steenkiste, Sjoerd and Kurach, Karol and Marinier, Rapha{\"e}l and Michalski, Marcin and Gelly, Sylvain},
  year={2019}
}

@inproceedings{deitke2023objaverse,
  title={Objaverse: A universe of annotated 3d objects},
  author={Deitke, Matt and Schwenk, Dustin and Salvador, Jordi and Weihs, Luca and Michel, Oscar and VanderBilt, Eli and Schmidt, Ludwig and Ehsani, Kiana and Kembhavi, Aniruddha and Farhadi, Ali},
  booktitle={Proceedings of the IEEE/CVF conference on computer vision and pattern recognition},
  pages={13142--13153},
  year={2023}
}

@article{comanici2025gemini,
  title={Gemini 2.5: Pushing the frontier with advanced reasoning, multimodality, long context, and next generation agentic capabilities},
  author={Comanici, Gheorghe and Bieber, Eric and Schaekermann, Mike and Pasupat, Ice and Sachdeva, Noveen and Dhillon, Inderjit and Blistein, Marcel and Ram, Ori and Zhang, Dan and Rosen, Evan and others},
  journal={arXiv preprint arXiv:2507.06261},
  year={2025}
}

@article{simeoni2025dinov3,
  title={Dinov3},
  author={Sim{\'e}oni, Oriane and Vo, Huy V and Seitzer, Maximilian and Baldassarre, Federico and Oquab, Maxime and Jose, Cijo and Khalidov, Vasil and Szafraniec, Marc and Yi, Seungeun and Ramamonjisoa, Micha{\"e}l and others},
  journal={arXiv preprint arXiv:2508.10104},
  year={2025}
}

@article{zhou2025learning,
  title={Learning 3d persistent embodied world models},
  author={Zhou, Siyuan and Du, Yilun and Yang, Yuncong and Han, Lei and Chen, Peihao and Yeung, Dit-Yan and Gan, Chuang},
  journal={arXiv preprint arXiv:2505.05495},
  year={2025}
}

@article{shen2026lyra,
  title={Lyra 2.0: Explorable Generative 3D Worlds},
  author={Shen, Tianchang and Bahmani, Sherwin and He, Kai and Srinivasan, Sangeetha Grama and Cao, Tianshi and Ren, Jiawei and Li, Ruilong and Wang, Zian and Sharp, Nicholas and Gojcic, Zan and others},
  journal={arXiv preprint arXiv:2604.13036},
  year={2026}
}

@article{li2025flashworld,
  title={FlashWorld: High-quality 3D Scene Generation within Seconds},
  author={Li, Xinyang and Wang, Tengfei and Gu, Zixiao and Zhang, Shengchuan and Guo, Chunchao and Cao, Liujuan},
  journal={arXiv preprint arXiv:2510.13678},
  year={2025}
}

@article{wang2026rein3d,
  title={Rein3D: Reinforced 3D Indoor Scene Generation with Panoramic Video Diffusion Models},
  author={Wang, Dehui and Xu, Congsheng and Wei, Rong and Shi, Yue and Chen, Shoufa and Luo, Dingxiang and Yang, Tianshuo and Yang, Xiaokang and Qin, Yusen and Tang, Rui and others},
  journal={arXiv preprint arXiv:2604.10578},
  year={2026}
}

@article{ye2026semgs,
  title={SemGS: Feed-Forward Semantic 3D Gaussian Splatting from Sparse Views for Generalizable Scene Understanding},
  author={Ye, Sheng and Dong, Zhen-Hui and Fan, Ruoyu and Lv, Tian and Liu, Yong-Jin},
  journal={arXiv preprint arXiv:2603.02548},
  year={2026}
}

@misc{yang2025qwen3technicalreport,
      title={Qwen3 Technical Report}, 
      author={An Yang and Anfeng Li and Baosong Yang and Beichen Zhang and Binyuan Hui and Bo Zheng and Bowen Yu and Chang Gao and Chengen Huang and Chenxu Lv and Chujie Zheng and Dayiheng Liu and Fan Zhou and Fei Huang and Feng Hu and Hao Ge and Haoran Wei and Huan Lin and Jialong Tang and Jian Yang and Jianhong Tu and Jianwei Zhang and Jianxin Yang and Jiaxi Yang and Jing Zhou and Jingren Zhou and Junyang Lin and Kai Dang and Keqin Bao and Kexin Yang and Le Yu and Lianghao Deng and Mei Li and Mingfeng Xue and Mingze Li and Pei Zhang and Peng Wang and Qin Zhu and Rui Men and Ruize Gao and Shixuan Liu and Shuang Luo and Tianhao Li and Tianyi Tang and Wenbiao Yin and Xingzhang Ren and Xinyu Wang and Xinyu Zhang and Xuancheng Ren and Yang Fan and Yang Su and Yichang Zhang and Yinger Zhang and Yu Wan and Yuqiong Liu and Zekun Wang and Zeyu Cui and Zhenru Zhang and Zhipeng Zhou and Zihan Qiu},
      year={2025},
      eprint={2505.09388},
      archivePrefix={arXiv},
      primaryClass={cs.CL},
      url={https://arxiv.org/abs/2505.09388}, 
}

\clearpage
\beginsupplement

\begin{center}
     \Large\textbf{Appendix}
\end{center}

\section{Extended Related Work} \label{appendix:related_work}

\begin{table*}[h]
\centering
\caption{Comparison of generative world model capabilities. We use \cmark\ to indicate the capability is supported and \xmark\ otherwise. ``Scene Mem.'' denotes memory of observed scene regions; ``2D Imag.'' denotes pixel-space multi-hypothesis imagination; ``3D Imag.'' denotes multi-hypothesis imagination of explicit 3D representations beyond observed geometry; ``Sequential'' denotes online updates from streaming observations; and ``Semantic'' denotes semantic features or labels.}
\vspace{-5pt}
\label{tab:relatedwork-compare}
\setlength{\tabcolsep}{5pt}
\small
\renewcommand{\arraystretch}{0.85}
\begin{tabular}{lccccc}
\toprule
\textbf{Models}
& \textbf{Scene Mem.} & \textbf{2D Imag.} & \textbf{3D Imag.} & \textbf{Sequential} & \textbf{Semantic} \\
\midrule
DFoT~\cite{song2025history}              & \xmark & \cmark & \xmark & \cmark & \xmark \\
NWM~\cite{bar2025navigation}             & \xmark & \cmark & \xmark & \cmark & \xmark \\
Persistent EWM~\cite{zhou2025learning}   & \cmark & \cmark & \xmark & \cmark & \xmark \\
GEN3C~\cite{ren2025gen3c}                & \cmark & \cmark & \xmark & \cmark & \xmark \\
ViewCrafter~\cite{yu2024viewcrafter}     & \cmark & \cmark & \xmark & \xmark & \xmark \\
TrajectoryCrafter~\cite{yu2025trajectorycrafter} & \cmark & \cmark & \xmark & \xmark & \xmark \\
MVSplat~\cite{chen2024mvsplat}           & \cmark & \xmark & \xmark & \xmark & \xmark \\
VGGT~\cite{wang2025vggt}                 & \cmark & \xmark & \xmark & \xmark & \xmark \\
CUT3R~\cite{wang2025continuous}          & \cmark & \xmark & \xmark & \cmark & \xmark \\
DFM~\cite{tewari2023diffusion}           & \cmark & \cmark & \xmark & \cmark & \xmark \\
Lyra 2.0~\cite{shen2026lyra}             & \cmark & \cmark & \cmark & \xmark & \xmark \\
Rein3D~\cite{wang2026rein3d}             & \cmark & \cmark & \cmark & \xmark & \xmark \\
FlashWorld~\cite{li2025flashworld}       & \xmark & \cmark & \cmark & \xmark & \xmark \\
SemGS~\cite{ye2026semgs}                 & \cmark & \xmark & \xmark & \xmark & \cmark \\
\textbf{3D-Belief}                       & \cmark & \cmark & \cmark & \cmark & \cmark \\
\bottomrule
\end{tabular}
\vspace{-10pt}
\end{table*}

\section{Diffusion Training Formulation} \label{appendix:diffusion}

The forward process is defined as:
\begin{equation}
    q(o^{\text{trgt}}_\tau \mid o^{\text{trgt}}_{\tau-1}) = \mathcal{N}(o^{\text{trgt}}_\tau ; \sqrt{1 - \beta_\tau} o^{\text{trgt}}_{\tau-1}, \beta_\tau I).
\end{equation}

In the reverse process, we reconstruct $o^{\text{trgt}}$ conditioned on $o^{\text{ctxt}}$ and camera parameters $\phi^{\text{trgt}}$:
\begin{align}
&p_\theta(o^{\text{trgt}}_{0:T} \mid o^{\text{ctxt}}; \phi^{\text{ctxt}}, \phi^{\text{trgt}})\\
    &=p(o^{\text{trgt}}_T) \prod_{\tau=0}^T p_\theta(o^{\text{trgt}}_{\tau-1} \mid o^{\text{trgt}}_\tau, o^{\text{ctxt}}; \phi^{\text{ctxt}}, \phi^{\text{trgt}}), 
\end{align}
where $p_\theta(o^{\text{trgt}}_{\tau-1} \mid o^{\text{trgt}}_\tau, o^{\text{ctxt}}; \phi^{\text{ctxt}}, \phi^{\text{trgt}})$ is implemented by first predicting an intermediate clean 3DGS scene $z_\tau$ and then mapping it to a denoised observation using the Gaussian Splatting rendering function \(\mathcal{G}(\cdot)\):
\begin{align}
    z_{\tau-1} &= \Phi_\theta(o^{\text{ctxt}}, o^{\text{trgt}}_\tau; \tau, \phi^{\text{ctxt}}, \phi^{\text{trgt}}), \label{eq:appendix_sample_z}\\
    \hat{o}^{\text{trgt}}_{\tau-1} &= \mathcal{G}(z_{\tau-1}, \phi^{\text{trgt}}), \label{eq:appendix_sample_o_hat}\\
    o^{\text{trgt}}_{\tau-1} &\sim \mathcal{N}(o^{\text{trgt}}_{\tau-1}; C_{\tau-1} \hat{o}^{\text{trgt}}_{\tau-1}, \hat{\beta}_\tau I). \label{eq:appendix_sample_o}
\end{align}

Here, $\hat{o}^{\text{trgt}}_{\tau-1}$ serves as an estimate of the clean observation. The constants $C_{\tau-1}$ and $\hat{\beta}_{\tau-1}$ are specifically chosen to ensure the noise at time $\tau-1$ aligns with the total noise introduced during the forward process. $\Phi_\theta$ is a neural network shown in Figure~\ref{fig:architecture_bottom}. During test time, the scene is generated by iteratively applying Eqs.~\eqref{eq:appendix_sample_z}-\eqref{eq:appendix_sample_o}, starting from an initial distribution $p(o^{\text{trgt}}_{\tau=T}) \sim \mathcal{N}(0, I)$. 

The model defines a generative process over the Gaussian primitives, formalized as:
\begin{equation}
    p_{\theta, \phi^{\text{trgt}}}(z_{0:T} \mid o^{\text{ctxt}}; \phi^{\text{ctxt}}) = \prod_{\tau=1}^T p_\theta(z_{\tau-1} \mid o^{\text{trgt}}_\tau, o^{\text{ctxt}}; \phi^{\text{ctxt}}, \phi^{\text{trgt}}).
\end{equation}

\section{Extended Results}
\label{extended_results}

\subsection{Extended Vision Results}

\subsubsection{More Qualitative Results} \label{appendix:qualitative}

\begin{figure}[b!]
  \centering
  \includegraphics[width=1.0\linewidth]{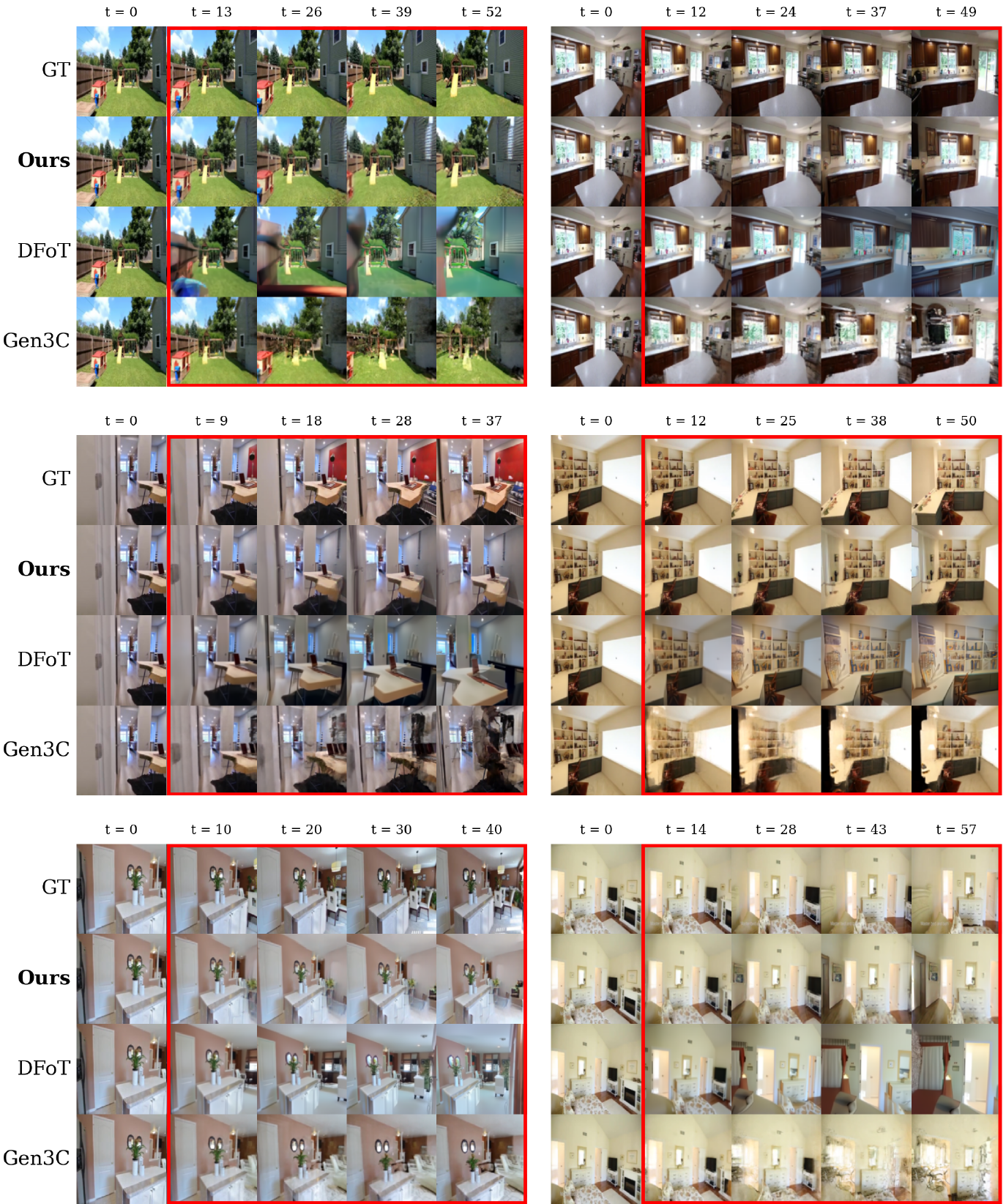}
  \vspace{4pt}

  \vspace{-6pt}
  \caption{Qualitative results on RE10K.}
  \label{fig:large_re10k}
  \vspace{-10pt}
\end{figure}

\begin{figure}[t!]
  \centering
    \includegraphics[width=1.0\linewidth]{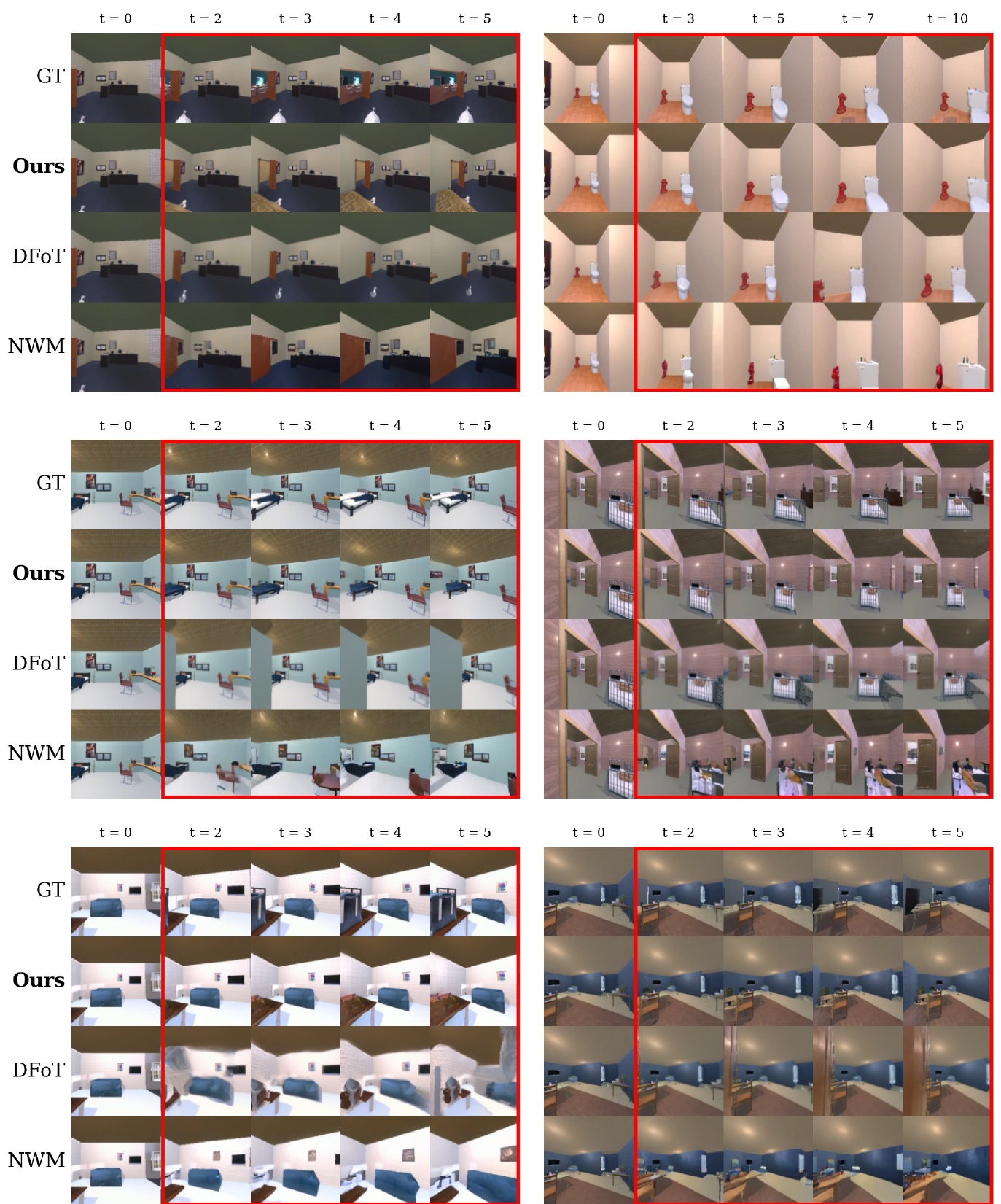}
    \vspace{-10pt}
    \caption{Qualitative results on AI2THOR.}
  \label{fig:large_ai2thor}
  \vspace{-10pt}
\end{figure}

We include more qualitative results in Figure~\ref{fig:large_re10k} and \ref{fig:large_ai2thor}.

\subsection{Extended Planning Results} \label{appendix:planning_results}

\subsubsection{Full Navigation Results}
We report the full object navigation results in Table~\ref{tab:fullobjsearch-results}, including additional baselines and inference speed comparisons.

\begin{table*}[t]
\centering
\caption{\textbf{Object navigation results in simulations and the real world.} 
Higher is better for SR, SPL, and SEL; lower is better for inference time and VLM tokens. 
Best results are highlighted in \textbf{bold}, and second-best simulation results are underlined.}
\vspace{-5pt}
\label{tab:fullobjsearch-results}
\scriptsize
\setlength{\tabcolsep}{4pt}
\renewcommand{\arraystretch}{1.15}

\resizebox{\linewidth}{!}{
\begin{tabular}{lccccc}
\toprule
\textbf{Model} &
\textbf{SR\% $\uparrow$} &
\textbf{SPL\% $\uparrow$} &
\textbf{SEL\% $\uparrow$} &
\makecell{\textbf{Inf.}\\\textbf{(s/step) $\downarrow$}} &
\makecell{\textbf{Token}\\\textbf{(/step) $\downarrow$}} \\
\midrule
\midrule

\multicolumn{6}{c}{\textbf{Simulations}} \\
\midrule
VGGT (w/ frontier)
& 27.50 & 25.82 & 22.66 & 25.13 & \textbf{0} \\
VGGT (w/ GPT-5m)
& 25.00 & 24.17 & 22.40 & 17.92 & 468.66 \\
VGGT (w/ Gemini 3.0)
& 25.00 & 24.10 & 22.66 & 18.87 & 1448.02 \\
DFoT-VGGT (w/ GPT-5m)
& 21.01 & 20.73 & 20.48 & 32.80 & 423.86 \\
DFoT-VGGT (w/ Gemini 3.0)
& 26.05 & 24.59 & 23.43 & 45.57 & 1210.08 \\
NWM-VGGT (w/ GPT-5m)
& 25.00 & 23.35 & 23.00 & 62.75 & 315.28 \\
NWM-VGGT (w/ Gemini 3.0)
& 25.00 & 23.46 & 20.76 & 54.41 & 552.12 \\
GPT-5m
& 23.33 & 17.66 & 20.19 & 11.10 & 6644.90 \\
Gemini 3.0
& \underline{45.00} & \underline{37.81} & \textbf{41.47} & 7.58 & 7512.70 \\
Qwen3-VL-8B-Instruct
& 18.33 & 14.08 & 16.94 & 2.30 & \underline{220.29} \\
SPOC
& 31.37 & 30.97 & 30.56 & \textbf{0.22} & \textbf{0} \\
\textbf{3D-Belief}
& \textbf{59.17} & \textbf{39.07} & \underline{40.24} & \underline{2.14} & \textbf{0} \\

\midrule
\midrule
\multicolumn{6}{c}{\textbf{Real World}} \\
\midrule
Gemini 3.0
& 23.08 & -- & 13.55 & 4.52 & 2317.09 \\
\textbf{3D-Belief}
& \textbf{55.56} & -- & \textbf{35.91} & \textbf{2.02} & \textbf{0} \\

\bottomrule
\end{tabular}
}
\vspace{-10pt}
\end{table*}

\subsubsection{Ablation Studies}\label{sec:app_ablation}
We report ablations for the 3D-Belief model in the simulated object navigation task in Table~\ref{tab:objsearch-ablations}.

\begin{table*}[h]
\centering
\caption{Ablation studies for simulated object navigation tasks.}
\vspace{-5pt}
\label{tab:objsearch-ablations}
\setlength{\tabcolsep}{1pt}
\begin{tabular}{lcccc}
\toprule
\textbf{Models} &
\textbf{SR\% $\uparrow$} &
\textbf{SPL\% $\uparrow$} &
\textbf{SEL\% $\uparrow$} &
\textbf{Token (/step) $\downarrow$} \\
\midrule
3D-Belief     & 45.83    & 36.43    & 32.99    & 0 \\
\quad w/o geometry    & 17.50     & 13.25    & 14.75   & 0 \\
\quad single hypothesis & 35.14 & 28.85 & 26.81 & 0 \\
\bottomrule
\end{tabular}
\vspace{-10pt}
\end{table*}

\subsubsection{Statistical Variation}
\label{appendix:variance}
In Table \ref{tab:objsearch-variance}, we present per-episode variability for simulated object navigation, comparing our 3D-Belief model with the strongest baseline, Gemini 3.0. We report 95\% confidence intervals computed over episodes, using Wilson binomial intervals for success rate and percentile bootstrap intervals for all other episode-level metrics.

\begin{table*}[h]
\centering
\caption{Episode-level performance for simulated object navigation. Values are means with 95\% confidence intervals in brackets.}
\vspace{-5pt}
\label{tab:objsearch-variance}
\setlength{\tabcolsep}{4pt}
\begin{tabular}{lccc}
\toprule
\textbf{Models} &
\textbf{SR\% $\uparrow$} &
\textbf{SPL\% $\uparrow$} &
\textbf{SEL\% $\uparrow$} \\
\midrule
Strongest baseline & $45.00\,[36.39, 53.92]$ & $37.81\,[29.55, 46.18]$ & $41.47\,[33.12, 50.02]$ \\
3D-Belief          & $59.17\,[50.22, 67.54]$ & $39.07\,[31.50, 46.87]$ & $40.24\,[32.70, 47.65]$ \\
\bottomrule
\end{tabular}
\vspace{-10pt}
\end{table*}

\subsubsection{Use VLMs for Semantics}
To study the impact of the built-in open-vocabulary semantics in 3D-Belief, we performed extra experiments with VLMs as the model semantic support. Results are reported in Table~\ref{tab:objsearch-extended}. We find that with VLMs, 3D-Belief model can perform better in navigation tasks, indicating that VLMs can help improve the semantic performance in terms of embodied decision making (e.g. selecting among different candidate paths). The performance of the built-in semantic module is largely affected by the CLIP-style vision-language alignment foundation models.

\begin{table*}[h]
\centering
\caption{Extended results with VLMs for semantic representations in simulated object navigation tasks.}
\vspace{-5pt}
\label{tab:objsearch-extended}
\setlength{\tabcolsep}{1pt}
\begin{tabular}{lcccc}
\toprule
\textbf{Models} &
\textbf{SR\% $\uparrow$} &
\textbf{SPL\% $\uparrow$} &
\textbf{SEL\% $\uparrow$} &
\textbf{Token (/step) $\downarrow$} \\
\midrule
3D-Belief     & 45.83    & 36.43    & 32.99    & 0 \\
\quad w/o semantics (w/ GPT-5m) & 49.58    & 40.10    & 41.01   & 79.42 \\
\quad w/o semantics (w/ Gemini 3) & 46.22    & 36.34    & 38.02  & 340.67 \\
\bottomrule
\end{tabular}
\vspace{-10pt}
\end{table*}

\subsection{Full Results on 3D-CORE} 
\subsubsection{Object Completion}
\label{appendix:objcomp_results}
We report the full object completion results with different visibilities in Table~\ref{tab:objcomp_full}.

\begin{table*}[h]
\centering
\caption{Results on Object Completion across different visibility.}
\begin{tabular}{llccccc}
\toprule
\textbf{Models} & \textbf{Visibility} & \textbf{BEV IoU}$\uparrow$ & \textbf{3D IoU}$\uparrow$ & \textbf{Chamfer}$\downarrow$ & \textbf{SigLIP}$\uparrow$ & \textbf{Recognition}$\uparrow$ \\
\midrule
\multirow{3}{*}{DFoT-VGGT}
& 0.05 & $0.110$ & $0.064$ & $2.681$ & $0.265$ & $0.126$ \\
& 0.55 & $0.362$ & $0.243$ & $0.830$ & $0.798$ & $0.767$ \\
& 0.95 & $0.372$ & $0.242$ & $0.189$ & $0.857$ & $0.838$ \\
\midrule
\multirow{3}{*}{\textbf{3D Belief}}
& 0.05 & $\mathbf{0.147}$ & $\mathbf{0.083}$ & $\mathbf{2.435}$ & $\mathbf{0.329}$ & $\mathbf{0.165}$ \\
& 0.55 & $\mathbf{0.484}$ & $\mathbf{0.318}$ & $\mathbf{0.216}$ & $\mathbf{0.855}$ & $\mathbf{0.930}$ \\
& 0.95 & $\mathbf{0.535}$ & $\mathbf{0.369}$ & $\mathbf{0.187}$ & $\mathbf{0.884}$ & $\mathbf{0.909}$ \\
\bottomrule
\end{tabular}
\label{tab:objcomp_full}
\end{table*}

\subsubsection{Room Completion}
We report room completion results with more metrics in Table~\ref{tab:roomcomp_full}.

\begin{table*}[h]
\centering
\caption{Results on Room Completion.}
\label{tab:roomcomp_full}
\resizebox{\linewidth}{!}{%
\begin{tabular}{lccccccc}
\toprule
\textbf{Models}
& \textbf{Obj. Precision $\uparrow$}
& \textbf{Obj. Recall $\uparrow$}
& \textbf{Obj. F1 $\uparrow$}
& \textbf{Occ. Acc. $\uparrow$}
& \textbf{IoU Free $\uparrow$}
& \textbf{IoU Occupied $\uparrow$}
& \textbf{Occ. IoU $\uparrow$} \\
\midrule
DFoT-VGGT
& 0.639
& \textbf{0.516 }
& 0.531 
& 0.252 
& 0.104 
& 0.115 
& 0.110  \\
\textbf{3D Belief}
& \textbf{0.678}
& 0.490
& \textbf{0.536}
& \textbf{0.900}
& \textbf{0.648}
& \textbf{0.235}
& \textbf{0.442} \\
\bottomrule
\end{tabular}%
}
\end{table*}

\subsection{Spatial Reasoning QA} \label{appendix:spatial_reasoning}

\subsubsection{Evaluation on SAT-Real}
To compare 3D-Belief with prior world models on the benefits they provide for reasoning about motion and space, we evaluate on the SAT-Real benchmark, which probes VLM reasoning by presenting two RGB images and asking binary questions over multiple subsets, namely Egocentric Movement (EM), Object Movement (OM), Egocentric-Allocentric Perspective Taking (Pers), Goal Aiming (GA), and Egocentric Action Consequence (EA).

\textbf{Baselines.}
We follow MindJourney~\cite{yang2025mindjourney} to perform test-time scaling with a world model: given the current observation, we use beam search to simulate plausible camera motions and generate additional visual evidence using the world model, which is then provided to the VLM as context for answering questions.
We compare 3D-Belief against two video world models, \textbf{SWM}~\cite{yang2025mindjourney} and \textbf{SVC}~\cite{zhou2025stable}, using the same inference protocol and budget (same beam width and rollout length).
All methods are paired with the same VLM backbone (Gemini-3.0), and differ only in the auxiliary world model used to generate the additional context.

\textbf{Results.}
Table~\ref{tab:satreal-result} shows that augmenting Gemini-3.0 with 3D-Belief yields the best overall performance (88.7\%), outperforming both the base VLM (85.3\%) and the best video-world-model baseline (86.7\%).
The gains are concentrated on subsets that require reasoning about camera motion and 3D space: 3D-Belief improves EM to 100.0\% (vs. 95.7\%), and substantially boosts EA (97.3\% vs. 83.8\%) and GA (94.1\% vs. 85.3\%).
These improvements are consistent with 3D-Belief providing an explicit and more geometrically grounded belief update, which better supports motion- and space-related inference under viewpoint changes.
Meanwhile, performance on OM and Pers is not the best among methods (e.g., OM 73.9\% vs.\ 78.3\%, Pers 75.8\% vs.\ 84.8\%), suggesting that fine-grained object-centric dynamics remain challenging and may require stronger object-level temporal modeling or higher-fidelity appearance tracking.

\begin{table}[h]
\centering
\setlength{\tabcolsep}{5pt}
\caption{Results on SAT-Real.}
\begin{tabular}{lcccccc}
\toprule
 & \multicolumn{6}{c}{\textbf{SAT Real}} \\
\cmidrule(lr){2-7}
 & \textbf{Avg} & \textbf{EM} & \textbf{OM} & \textbf{EA} & \textbf{GA} & \textbf{Pers} \\
\midrule
\textbf{Gemini-3.0} & 85.3 & 95.7 & \textbf{78.3} & 83.8 & 85.3 & \textbf{84.8} \\
\quad + SWM & 86.7 & 95.0 & 70.0 & 94.1 & 89.7 & 81.3 \\
\quad + SVC & 86.0 & 95.7 & \textbf{78.3} & 86.5 & \textbf{94.1} & 75.8 \\
\quad + \textbf{3D Belief} & \textbf{88.7} & \textbf{100.0} & 73.9 & \textbf{97.3} & \textbf{94.1} & 75.8 \\
\bottomrule
\end{tabular}
\label{tab:satreal-result}
\end{table}

\section{Implementation Details} \label{appendix:implementation}
\subsection{Model Training}

\subsubsection{Datasets}
We train our model on a composite dataset, which consists of SPOC \cite{ehsani2024spoc}, RealEstate10K \cite{zhou2018stereo}, DL3DV \cite{ling2024dl3dv}, and Habitat-Matterport 3D \cite{ramakrishnan2021hm3d}. At each step, we randomly select a subset with equal weights and sample frames. Each sample includes the context view and the target view with randomly spaced frames. To align the datasets, we set the minimum interval to 2. The maximum intervals for SPOC, RealEstate10K, DL3DV, and Habitat-Matterport 3D are 10, 100, 10, and 15, respectively.

\subsubsection{Training Settings}
\textbf{3D-Belief} The U-ViT backbone \cite{hoogeboom2023simple, hoogeboom2024simpler} used to encode input images and the depth predictor used to estimate Gaussian primitives are initialized from pretrained weights: the DFoT encoder \cite{song2025history} and the MVSplat multi-view depth predictor \cite{chen2024mvsplat}, respectively, both pretrained on RealEstate10K \cite{zhou2018stereo}. We use AdamW optimizer with a learning rate of $2 \times10^{-5}$ and weight decay of 0.001 for training, applying linear warm-up in the beginning 10K steps followed by cosine decay. The model can still be trained effectively and converge even with a small batch size. We set the batch size to 1 and trained for a total of 250K steps on a single H100 80GB chip.
We follow DFM \cite{tewari2023diffusion} to apply loss on the denoised target view image. We calculate L1 loss and LPIPS loss with a weight of 1.0 between the predicted image and the ground-truth. Although deep supervision is not employed, the model incorporates a depth smoothness loss with a weight of 0.1 as a regularization term. Additionally, the semantic loss based on CLIP \cite{ilharco2021openclip} and DINOv3 \cite{simeoni2025dinov3} is calculated separately, with a weight of 0.1. Besides the target view, we also compute all the aforementioned loss for intermediate frames to enhance consistency. To further enhance the model's capabilities, we align the prediction and ground truth at the feature level with a weight of 5.0, while simultaneously aligning the prediction features using the pre-trained VGGT \cite{wang2025vggt} model with a weight of 2.0. Training is completed by performing a weighted sum of all losses.

\textbf{DFoT}
We use the DFoT model \cite{song2025history} trained on RealEstate10K \cite{zhou2018stereo} as the base model, and finetune it on the SPOC \cite{ehsani2024spoc} dataset with light horizontal flip augmentation. Training uses AdamW with learning rate $1 \times10^{-5}$, weight decay 0.01, per-GPU batch size 4 with gradient accumulation of 2, and runs for 48 epochs on 4 $\times$ H100 80GB chips. The diffusion process follows a cosine schedule with sigmoid loss weighting.

\textbf{NWM} We use the official pretrained NWM \cite{bar2025navigation} CDiT/XL model, which covers four robotics datasets (SCAND \cite{karnan2022scand}, TartanDrive \cite{triest2022tartandrive}, RECON \cite{shah2021recon}, and HuRoN \cite{hirose2023sacson}) and Ego4D \cite{grauman2022ego4d} videos. Then, we finetune it on SPOC \cite{ehsani2024spoc} dataset using AdamW optimizer with a learning rate of $8 \times10^{-5}$ and a batch size of 16, and a total of 200K training steps. Training is done on a single H100 80GB chip.

\textbf{VGGT} We utilize the official pretrained VGGT-1B model \cite{wang2025vggt} with both camera and depth heads enabled. We finetune it on the SPOC \cite{ehsani2024spoc} dataset on $238 \times 238$ inputs (patch size 14) using AdamW (learning rate $1 \times 10^-5$, weight decay 0.05) for 60 epochs with an effective batch size of 96. Training is done on 4 $\times$ GH200 96GB chips.

\textbf{SPOC} We finetune the pretrained SigLIP-ViTb-3 CHORES-Nav model \cite{ehsani2024spoc} (an EarlyFusionCnnTransformer policy) on the SPOC \cite{ehsani2024spoc} RERENDERED mixture, which combines all four ObjectNav task types (ObjectNavType, ObjectNavRoom, ObjectNavRelAttribute, and ObjectNavLocalRef). The policy consumes raw navigation and manipulation camera streams together with the agent's last-action and in-hand-object signals. Training is performed on 4 $\times$ H100 80GB chips with a per-GPU batch size of 6 and a sliding window of 50 steps, using a learning rate of $2 \times 10^{-5}$ and mixed-precision (fp16) for 50 epochs over a 100{,}000-sample subset of the dataset.

\subsection{Planning} \label{appendix:planning}

\subsubsection{Object Navigation Tasks}

We provide additional implementation details for the object navigation experiments. At each step, the agent updates its 3D belief $z^t$ from the latest egocentric RGB stream and poses. A frontier goal sampler then proposes the next possible waypoint goals, computes candidate paths with an A* planner, and performs mental simulation by rendering imagined observations along each path from $z^t$. These rollouts are scored by goal progress and information gain via semantic-map queries. The agent executes a short prefix of the best plan and repeats.

We describe each key component below.

\textbf{Waypoint Sampling.} Planning directly to the final goal is difficult under partial observability, since large portions of the environment are unseen and therefore uncertain. We instead plan an intermediate waypoint at each step. Waypoints are selected in two cases. If the final goal (a target object or a target location) has already appeared in the observed history $o^{1:t}$, we set the waypoint to the estimated goal position by querying the semantic map. Otherwise, we sample a small set of candidate waypoints in currently unobserved regions within an exploration radius centered at the agent’s current position.

\textbf{Simulating and Evaluating the Paths.} For each of the waypoints, we can sample a path and simulate the agent's imagined observations along that path using the predicted scene representation $z^t$. These rollouts estimate how informative or goal-directed a path is. For example, whether the target object becomes visible in the imagined views or whether the rendered semantics indicate proximity to the goal. Specifically, we score each path by querying the rendered semantic feature maps to assess progress and promise from the imagined observations. If the semantics map indicates that it is likely that the object is on the imagined path, then we will score the path highly.

\textbf{Execution.} We execute the first $T$ steps of the highest-scoring plan, then re-estimate the scene representation from the newly acquired observations and repeat the procedure iteratively until the goal is reached.

In all planning experiments, we use 3 hypotheses for 3D-Belief sampling at each decision step. The results for the 3D-Belief model and SPOC are obtained using a single H100 80GB chip for inference, while other evaluations rely on a single A100 80GB chip.

\subsubsection{Real-World Experiments} \label{appendix:real}

\begin{figure}[t!]
  \centering
    \includegraphics[width=0.7\linewidth]{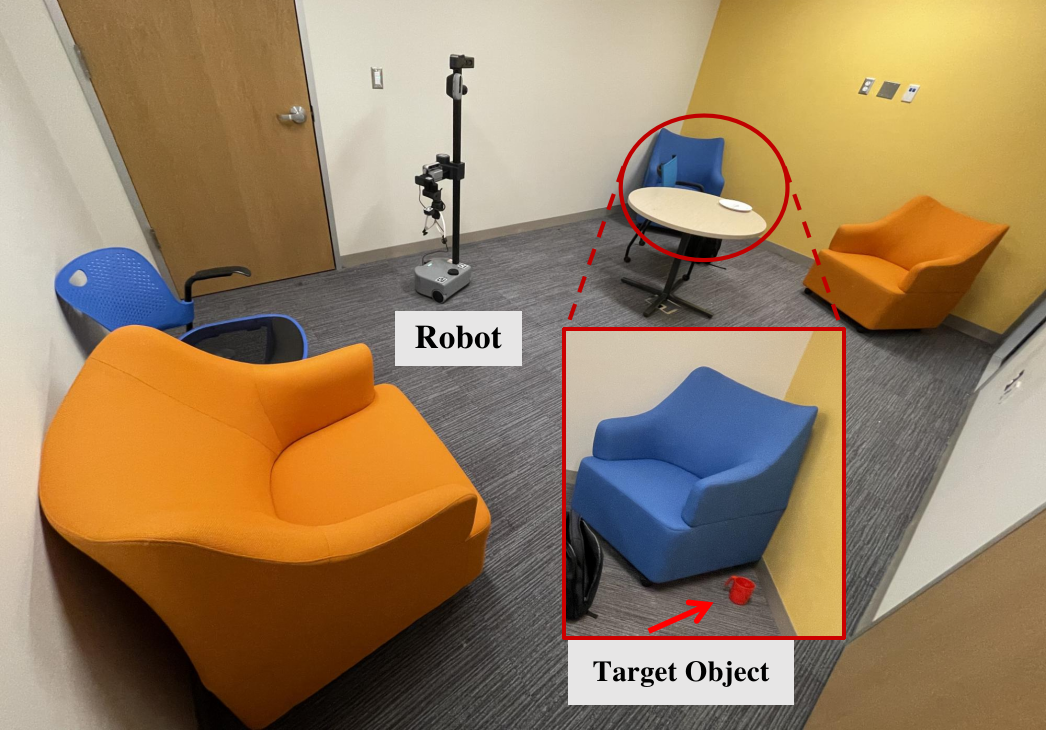}
  \caption{Example setup for the real-world object navigation tasks with 3D-Belief. The small image shows a zoom-in view near the blue couch with a target object hidden.}
  \label{fig:real_setup}
  \vspace{-15pt}
\end{figure}

For the object navigation task, we select 5 target objects and randomize each episode by (1) sampling a target object and (2) initializing the robot from one of 3 predefined starting locations to induce different exploration requirements. An example setup is shown in Fig.~\ref{fig:real_setup}.

We use the Stretch 3 Mobile Manipulator (Hello Robot) as our real-world embodied platform. The 3D-Belief model and all baselines run on an external GPU client (an NVIDIA RTX 4090 desktop), and the robot communicates with the client via ROS~2. 

We mount an RGB-D camera (\textit{Intel RealSense} D455) on the robot and use the RGB stream as the visual input. The robot's pose is estimated using onboard wheel-encoder odometry. Together, the RGB observations and associated poses form an egocentric stream that serves as input to our models. Note that 3D-Belief uses only RGB at inference time. However, the real-world vision datasets used for training provide camera poses from \textit{Structure-from-Motion (SfM)}, which are defined only up to an unknown global scale; as a result, the model's depth predictions are not inherently metric. To recover metric-consistent geometry for planning, we align the predicted depth map $\hat{d}$ to the sensed depth $d$ by estimating a per-sequence scalar scale $s$ via robust regression, and apply this scale to the depth maps produced by the Multi-view Depth Predictor (Sec.~\ref{sec:architecture}).

\subsection{Reasoning}
\subsubsection{3D-CORE Metrics} \label{appendix:metrics}
For \emph{Object Completion}, BEV IoU, 3D IoU, and Chamfer distance are calculated by the extracted point cloud of the imagined object (segmented by Gemini 2.5 \cite{comanici2025gemini}) and the GT point cloud (segmented by GT masks). SigLIP is calculated by comparing the predicted and GT views at the same camera poses. Recognition is defined by the success rate of target object recognition by VLMs in the imagined views.

For \emph{Room Completion}, Obj. F1 is calculated by comparing a list of VLM (Gemini 2.5 \cite{comanici2025gemini}) recognized objects in imagined views, with the GT object list in that view. Occ. Acc. is defined by the accuracy on cells where both gt and pred are known (0 or 1). IoU Free is the IoU for free cells (class 0), IoU Occupied is the IoU for occupied cells (class 1), Occ. IoU is the average of the previous two. For \emph{Object Permanence}, LPIPS and SigLIP are calculated using the first and the last rendered image, whose camera poses are the same. All reasoning experiments utilize a single A100 80GB chip for inference.

\section{Discussion}

\textbf{How does 3D imagination affect 2D rendering quality?} 

While 3D-Belief is designed for explicit 3D reasoning, it also improves \emph{2D} rendering quality as shown in Section~\ref{exp:vision}. Because predictions are structured in an explicit 3D scene, generated frames are constrained by coherent geometry instead of being synthesized purely in pixel space. This acts as a strong prior for cross-view consistency: once surfaces and free space are established in 3D, new viewpoints are rendered by projection, reducing viewpoint ambiguity and limiting texture drift or identity switches over long horizons. This yields better perceptual/distributional metrics (lower LPIPS/FID/FVD) and higher PSNR/SSIM. Qualitatively (Appendix~\ref{appendix:qualitative}), 2D-only baselines often drift under large camera motion, whereas 3D-Belief better preserves rigid structure and object permanence across revisits.

\textbf{How do the key 3D belief model capacities affect the embodied task performance?}

\begin{figure}[t]
  \centering
    \includegraphics[width=0.8\linewidth]{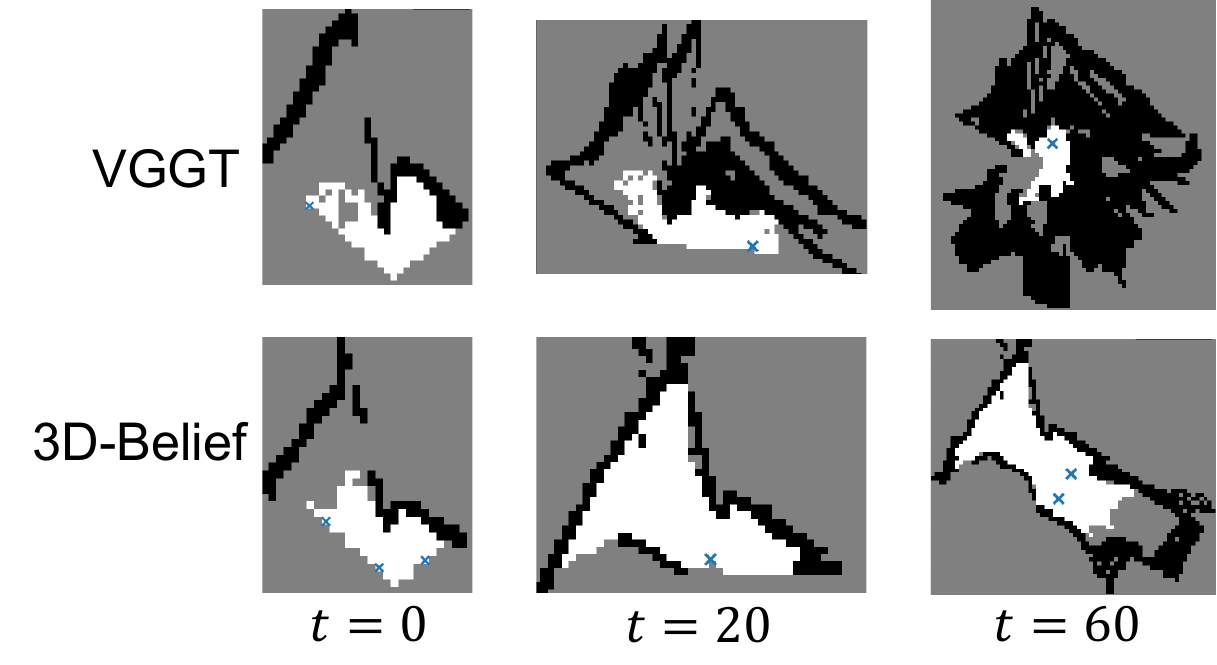}
    \vspace{-5pt}
  \caption{Comparison of inferred occupancy map generated by 3D-Belief and VGGT (w/ frontier) over time in the same task.}
  \label{fig:long_horizon_occ}
  \vspace{-15pt}
\end{figure}

A key advantage of 3D-Belief is maintaining a spatially consistent scene memory over long rollouts. In Figure~\ref{fig:long_horizon_occ}, VGGT-based occupancy quickly accumulates duplicated structure, and maps often break afterwards, trapping the agent. In contrast, 3D-Belief remains coherent and steadily expands over much longer horizons, supporting continued exploration and replanning. This failure mode is worse for DFoT-VGGT and NWM-VGGT, where both lifting and video prediction drift under long horizons, corrupting the 3D cache and degrading planning \cite{xiao2025worldmem, melnik2024video, oshima2025worldpack}.

Table~\ref{tab:objsearch-ablations} in Appendix~\ref{sec:app_ablation} further shows that removing multi-hypothesis sampling causes a performance drop, indicating that maintaining multiple plausible beliefs is important for robust decision making. Together with internal semantic prediction, this lets the agent compare candidate actions conditioned on task or language specifications, rather than relying mainly on waypoint sampling. It also reduces deployment cost: compared to attaching an external VLM for semantics, which adds token and compute overhead (Table~\ref{tab:objsearch-results}), 3D-Belief provides semantics within the model.

\section{Broader Impacts} \label{appendix:impacts}
3D-Belief aims to improve the ability of embodied agents to reason about partially observed environments by maintaining and updating explicit 3D beliefs over both observed and unobserved space. It has potential benefits for assistive robotics, object search, warehouse inspection, and navigation in areas like homes or hospitals. By representing uncertainty explicitly and supporting multi-hypothesis reasoning, it may help agents explore more efficiently and make better-informed decisions.

However, predicting unobserved regions also introduces risks. Plausible but incorrect beliefs could lead to unsafe robot behavior if treated as facts. In indoor settings, explicit 3D scene representations may also capture sensitive information about private spaces and personal belonging, raising privacy and surveillance concerns. This work should therefore be viewed as a research step rather than a deployment-ready system. Practical deployments should include conservative planning, collision checking, privacy-preserving data handling, and clear separation between observed evidence and imagined scene completions.

\section{Asset Licenses}
\label{appendix:licenses}
This section documents the licenses and terms of use for existing assets used in
the paper. We do not claim ownership of these
third-party assets, and we comply with their corresponding license and usage
requirements.

\begin{table}[h]
\centering
\small
\caption{Existing assets used in this work and their licenses or terms of use.}
\label{tab:asset_licenses}
\setlength{\tabcolsep}{4pt}
\begin{tabularx}{\linewidth}{@{}>{\raggedright\arraybackslash}p{0.43\linewidth}>{\raggedright\arraybackslash}X@{}}
\toprule
\textbf{Asset} & \textbf{License / Terms} \\
\midrule
AI2-THOR \cite{kolve2017ai2} & Apache-2.0 \\
ProcTHOR / ProcTHOR-10K \cite{NEURIPS2022_27c546ab} & Apache-2.0 \\
Objaverse \cite{deitke2023objaverse} & ODC-By v1.0 dataset license; per-object Creative Commons licenses \\
RealEstate10K \cite{zhou2018stereo} & CC-BY 4.0 \\
OpenCLIP \cite{ilharco2021openclip} & MIT \\
DINOv3 \cite{simeoni2025dinov3} & DINOv3 License \\
DFoT \cite{song2025history} & MIT \\
Navigation World Models (NWM) \cite{bar2025navigation} & CC-BY-NC 4.0 \\
VGGT \cite{wang2025vggt} & Meta Research Materials License \\
Qwen3-VL-8B-Instruct \cite{yang2025qwen3technicalreport} & Apache-2.0 \\
\bottomrule
\end{tabularx}
\end{table}

\end{document}